\documentclass[conference]{IEEEtran}
\IEEEoverridecommandlockouts
\usepackage{cite}
\usepackage{amsmath,amssymb,amsfonts}
\usepackage{algorithm}
\usepackage{algpseudocode}
\usepackage{array}
\usepackage{booktabs}
\usepackage{multirow}
\usepackage{graphicx}
\usepackage{svg}
\usepackage{textcomp}
\usepackage{float}
\usepackage[table]{xcolor}
\usepackage{xcolor}
\usepackage{comment}
\usepackage{makecell}
\usepackage[dvipsnames]{xcolor}
\usepackage[hyphens]{url}
\usepackage{hyperref}

\def\BibTeX{{\rm B\kern-.05em{\sc i\kern-.025em b}\kern-.08em
    T\kern-.1667em\lower.7ex\hbox{E}\kern-.125emX}}
\begin{document}

\newcommand{\CRH}[1]{\textcolor{red}{\textbf{Chaozhe:} #1\textbf{}}}

\newcommand{\HX}[1]{\textcolor{blue}{\textbf{HX:} #1\textbf{}}}

\newcommand{\RS}[1]{\textcolor{Green}{\textbf{Rishabh:} #1\textbf{}}}

\newcommand{\YR}[1]{\textcolor{Green}{\textbf{Yamini:} #1\textbf{}}}

\newcommand{\SSS}[1]{\textcolor{Magenta}{\textbf{Swarat:} #1\textbf{}}}

\title{Infrastructure-Guided Connectivity-Enhanced \\
Road Crack Detection and Estimation
}

\author{\IEEEauthorblockN{Haosong Xiao\textsuperscript{*}, Yamini Ramesh\textsuperscript{**}, Rishabh Shukla\textsuperscript{**},  
Swarat Sarkar \textsuperscript{**} and Chaozhe R. He\textsuperscript{*}}
\IEEEauthorblockA{
\textit{\textsuperscript{*}Department of Mechanical and Aerospace Engineering}\\
\textit{\textsuperscript{**}School of Engineering and Applied Science}\\
\textit{University at Buffalo (SUNY)}, Buffalo, New York, USA \\
\{haosongx, yaminira, rshukla3, swaratsa, chaozheh\}@buffalo.edu 
}
\thanks{The source code and data is available at

\url{https://github.com/CHELabUB/cav_road_crack_detection}
}
}

\maketitle
\thispagestyle{plain}
\pagestyle{plain}

\begin{abstract}
In this paper, we report the world's first infrastructure-guided communication-enhanced road crack detection pipeline that is effective and implementable on passenger vehicles. 
We first design a customized communication protocol to transmit the region of interest from the infrastructure to the vehicle. 
With proper camera image processing (e.g., dynamic cropping and frame selection), the focused images are provided to the crack detection model. 
Leveraging state-of-the-art crack detection model backbones and a carefully prepared dataset comprising a forward-facing view with a crack, we train the model to improve crack-detection performance. 
We demonstrate the full detection pipeline on an experimental vehicle platform, showcase the detection effectiveness, and project future research directions.
\end{abstract}

\section{Introduction}
Infrastructure maintenance will cost trillions of US dollars in investment between 2024 and 2033 \cite{ASCE_IRC_2025}. 
Among various categories of infrastructure, bridges, and more specifically, bridge decks face a significant funding gap for maintenance. 
Bridge decks account for a significant portion of overall bridge maintenance costs and directly affect safety and user experience \cite{Kong2022bridgeDeckDeterioration,FHWA2024bridgeDeckDeterioration, Ibrahim2024concreteBridgeDeckDeterioration}. 
It is estimated that 50\% to 85\% of total bridge maintenance costs are attributed to deck maintenance \cite{Omar2022ReviewConcreteBridgeDecks}. 
Deterioration mechanisms such as reflective cracking, joint spalling, and chloride-induced corrosion are well documented, particularly in precast decks with multiple panel joints \cite{Tawadrous2019PrecastConcreteBridgeDeck, Garber2021FDPCDeckPanels}. 
Early detection of these defects is crucial: once rebar corrosion and delamination advance, repairs become far more expensive and disruptive. 
Visible signs of damage from the surface, such as joint cracking, grout spalling, or reflective cracking, can indicate underlying issues, and could lead to structural degradation if unchecked. 
And overly aggressive cracking would require repairs which could lead to road closure, disrupting traffic \cite{FHWA2022DeckPanels}. 

Federal regulations mandate routine inspections every 24 months to ensure safety. 
Manual inspections remain the predominant means of inspection for professional tasks requiring detailed training \cite{AASHTO2011bridgeInspection}, leading to steep labor costs, traffic disruptions, and safety risks \cite{AASHTO2019dronesForInfrastructure}. 
Therefore, developing effective and efficient crack-detection approaches for the bridge surface inspection, or more general road surface crack detection, is crucial for maintaining bridge health in an efficient and economical manner. 
While remote sensing systems using drones and dedicated inspection vehicles are commercially available, their operating costs remain high, making continuous monitoring uneconomical \cite{AlynixDeckerAssessment, NEXCODeckTopScanningSystem}. 
Currently, there is a lack of cost-effective, automated inspection systems capable of detecting damage between routine inspections, particularly early signs of degradation. 
Developing such a system would enable continuous monitoring, reducing the risk of undetected damage and enhancing the overall durability of precast bridge components.

With the emerging vehicular technologies in sensing and connectivity, modern vehicles are increasingly capable of perceiving road conditions. 
Harnesses crowd sensing capability brought by road vehicles equipped with front-view cameras could be used to build a cost-effective, continuous bridge surface crack inspection system.
Nationwide deployment plan of Vehicle-to-everything (V2X) devices \cite{USDOT2024V2XDeployment} facilitates the feasibility of effective crowd sensing, further enabling such system.
Built on these new technologies, in this paper, we report our efforts towards such inspection system by making the following contributions:
\begin{enumerate}
\item We establish the world's first infrastructure-guided connectivity-enhanced crack detection framework that leverages all road vehicles for road surface inspections
\item We provide a detailed design recipe for communication protocol design, crack detection model tuning, camera calibration, and crack calculation.
\item We release the first dataset of surface cracks from the vehicle's front-view camera.
\item We implement and demonstrate the designed crack detection framework on a real-vehicle platform.
\end{enumerate}

\section{Background}\label{sec:background}
In this section, we first survey related work in the field. 
With a motivating example, we then highlight the limitations of existing work and preview the design in this work.

\subsection{Related Work}
Prior crack detection research can be divided into two categories. 
One focuses on improving the crack recognition accuracy by advancing the detection and segmentation models. 
Early approaches primarily relied on low-level image intensity analysis, using statistical analysis of pixels' surrounding intensity difference to recognize cracks from the pavement surface \cite{kirschke1992histogram}. 
Kernel-based edge detectors, such as Sobel and Canny operators, were also widely used \cite{abdel2003analysis,ayenu2008evaluating}.
These methods are sensitive to surface texture and noise due to lighting conditions. 
More recent techniques, including shadow moving and tensor voting, are used to reduce the impact of image noise \cite{zou2012cracktree}. 
The extensive use of parameters increase the effort of hand-tuning and limits generalization capabilities. 
These limitations can be mitigated by adopting learning-based approaches, which train neural networks on annotated pavement images to learn different crack features, thereby improving the robustness of the detection performance under illumination variation and surface texture variations \cite{zou2018deepcrack, chen2022refined, li2024cracktinynet}. 
 
The other focuses on implementing detection models on existing and purposefully developed unmanned aerial and ground vehicles (UAVs and UGVs) to streamline the detection process. 
Prior works using UAVs \cite{pan2018detection, chen2024pavement} are constrained by the limited battery life, and the hidden trade-off between flight height and image resolution. 
A purposefully designed platform combining UGV and robotic arm used in \cite{song2025robust} demonstrates strong performance detection performance on daily used traffic corners, but it also raises concern on disrupting traffic during regular detection task. 
Using a designated road vehicle with detection externally mounted cameras, as shown in \cite{wang2024measurement}, can mitigate the disturbance to traffic. 
Yet its detection range may be limited due to a fixed down-view camera angle and potentially strong requirements on the vehicle's maneuver (e.g., path and speed) for designated crack locations.
While front-view camera-based \cite{anand2018crack} and modification-free vehicle-based \cite{kortmann2022watch} platforms demonstrate the potential of autonomous vehicles for crack detection, the crack detection only focused on object detection rather than length estimation making it insufficient for bridge inspection usage. 

Modern vehicles with sensors such as front-view cameras could continuously monitor the road, allowing vehicles to assess the status of the infrastructure using vision. 
Existing work along this direction though focuses only on benefiting the vehicle, e.g., estimating road roughness and friction coefficient and detecting pavement cracks and potholes to improve ride comfort \cite{NexteerRoadSurfaceDetection,MercedesRoadSurfaceDetection}. 
Opportunities to extend such capability to improve infrastructure durability (e.g., bridge) are unexplored \cite{Ranyal2022SmartPhoneRoadConditionMonitoring}. 
One obstacle is enabling direct information sharing between vehicles and infrastructure regarding inspection results acquired by road vehicles. With the increase level of connectivity \cite{USDOT2024V2XDeployment} in transportation system, such obstacle can be overcome.
Yet challenges remains on how to effectively adapt vehicle centric sensors and algorithm design towards infrastructure inspection applications.   

Overall, there remains a gap in developing inspection approaches that enable continuous crack detection on pavement without disrupting traffic or limiting operational time, while maintaining robust detection accuracy.

\subsection{Motivating Connectivity and Dynamic Cropping}
\label{sec: motivation}

\begin{figure}[t]
\centering

\includegraphics[width=0.48\textwidth]{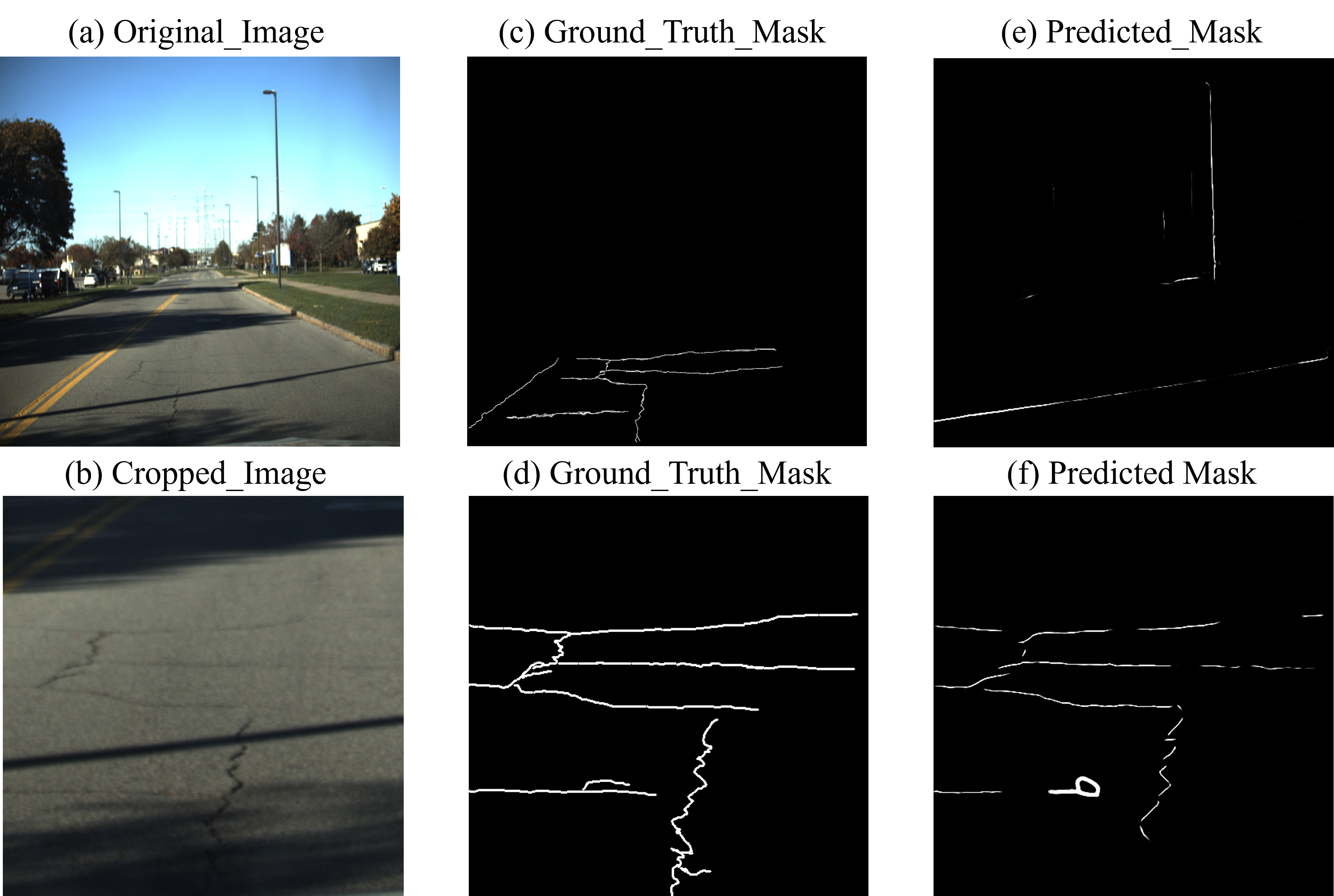}
\caption{Dynamic cropping impact: (a) Original full-frame image (2064×1544), (b) Cropped 512×512 region, (c) Ground truth mask of the original frame, (d) Corresponding ground truth of the cropped frame, (e) Predicted mask showing false positives from background noise, (f) Cleaner prediction with precision improvement from 0.20 to 0.70. 
}
\label{fig:cropping_benefit}
\end{figure}

\begin{table}[t]
\centering
\caption{Detection Metrics: Full Frame vs. Cropped}
\label{tab:cropping_metrics}
\small
\begin{tabular}{|l|c|c|c|}
\hline
\textbf{Metric} & \textbf{Full Frame} & \textbf{Cropped} & \textbf{Improvement} \\
\hline
ODS F1 & 0.2251 & 0.5288 & +135\% \\
\hline
Precision & 0.2009 & 0.6966 & +247\% \\
\hline
Recall & 0.3554 & 0.4780 & +34\% \\
\hline
OIS F1 & 0.2688 & 0.5336 & +99\% \\
\hline
AP & 0.0382 & 0.3521 & +822\% \\
\hline
\end{tabular}
\end{table}
\begin{figure*}[!t]
    \centering
    \includegraphics[width=\textwidth]{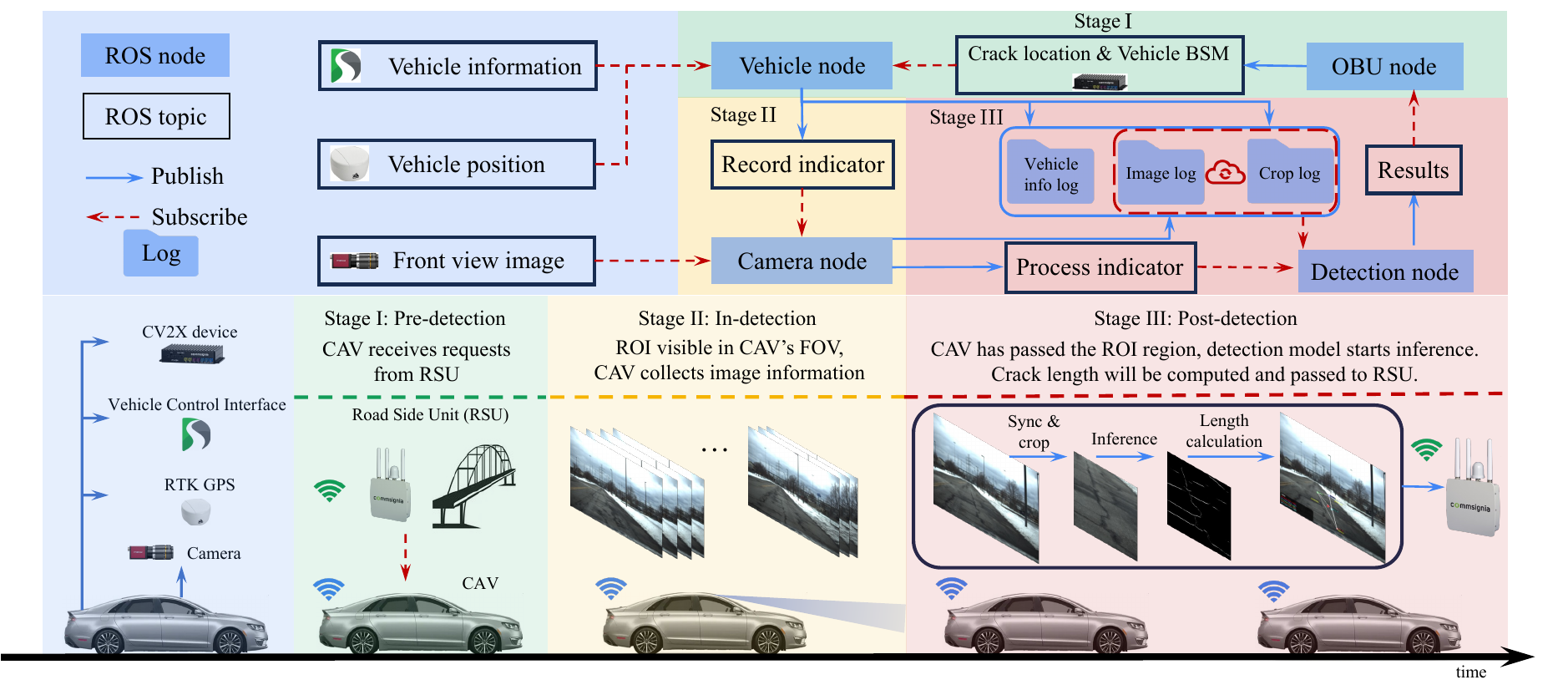}
    \caption{Design overview of the connectivity-enhanced crack detection pipeline.}
    \label{fig:design_overview}
\end{figure*}

Despite extensive research on automated crack detection~\cite{ma2025self}, benchmarks are primarily reported on limited datasets, and real-world operational performance remains limited. 
Existing methods often excel on curated datasets but struggle with the variable conditions encountered in real-world deployment. 
Real-world crack detection faces a fundamental challenge: full-frame processing of high-resolution front-view camera images contains substantial irrelevant background information that degrades model performance. 
To showcase this limitation and motivate our designs, we ran the developed crack detection model (details to be provided in Section \ref{sec:crack_detection_model}) to both full frame and properly cropped image. 
We evaluate detection performance using standard segmentation metrics, including Optimal Dataset Scale F1-score (ODS F1), Optimal Image Scale F1-score (OIS F1), precision, recall, and Average Precision (AP), comparing full 2064$\times$1544 frames against cropped regions. 
ODS F1 and OIS F1 denote the best F1-score over a fixed dataset-wide threshold and per-image adaptive thresholds, respectively, while AP summarizes precision–recall performance across all thresholds. 
As shown in Fig.~\ref{fig:cropping_benefit}, full-frame inference yields low precision (0.20) and ODS F1 (0.23), as background textures, shadows, and road markings introduce false positives. 

Motivated by this finding, we design a dynamic cropping algorithm in Section~\ref{sec:cropping_window} that captures such region of interest (ROI) and significantly reduces background interference. 
The algorithm is guided by infrastructure through connectivity, to compute an optimal cropping window centered on the target crack using the coordinate transformation pipeline described. 
Results in Table~\ref{tab:cropping_metrics} show improvements of 247\% in precision (from 0.20 to 0.70) and 135\% in ODS F1 (from 0.23 to 0.53), highlighting the effectiveness of the proposed approach. 

\subsection{Design Overview}
Our connectivity-guided crack detection pipeline, as shown in Fig.~\ref{fig:design_overview}, provides a novel, comprehensive, and extensible framework for autonomous road surface inspection. 
The pipeline consists of three stages. 
1) Pre-detection stage where communication happens between the infrastructure (e.g. bridge) and the vehicles via the cellular-vehicle-to-everything (C-V2X) protocol \cite{Qualcomm2019CV2XOverview} (c.f., Stage I in Fig.~\ref{fig:design_overview}). 
2) In-detection stage: an adjustable detection range is determined ensures optimal crack resolution using real-time onboard vehicle sensor data (c.f., Stage II in Fig.~\ref{fig:design_overview}).
3) Post-Detection stage: an improved crack detection model operates on a crack-centered cropped image from the synchronization mechanism ((c.f., Stage III in Fig.~\ref{fig:design_overview}) and delivers detection results.

The remainder of this paper elaborates on the details of various components: the mathematical formulation enabling accurate crop localization and subsequent crack length estimation in Section \ref{sec:CV2X_based_selection}; model training in Section \ref{sec:crack_detection_model}; full pipeline implementation and performance on a connected automated vehicle (CAV) platform in Section~\ref{sec:pipeline}.

\section{CV2X-based view selection}
\label{sec:CV2X_based_selection}
In this section, we first describe the algorithm that determine the focus area around the target crack on the road surface. 
Then we derive the crack length estimation algorithm based on vehicle location and target crack location. 
These two algorithms are crucial parts of Stage II and III in Fig.~\ref{fig:design_overview}

\subsection{Dynamic Cropping Window Design}\label{sec:cropping_window}
We assume that infrastructure has a known region of interest (ROI), whose center may contains cracks that critical to structural health. 
The vehicle receives the GPS coordinates of the region of interest from infrastructure through connectivity. 
Given the GPS coordinates of ROI, along with the vehicle's coordinate acquired from GPS (e.g., RTK GPS), we can convert the crack's relative position with respect to the vehicle into local east-north-up (ENU) coordinates, and eventually acquire the relative position of the crack with respect to the camera lens denoted as $\mathbf{p}'_{\rm crack2C}$.

\begin{figure}[H]
    \centering
    \includegraphics[width=0.5\textwidth]{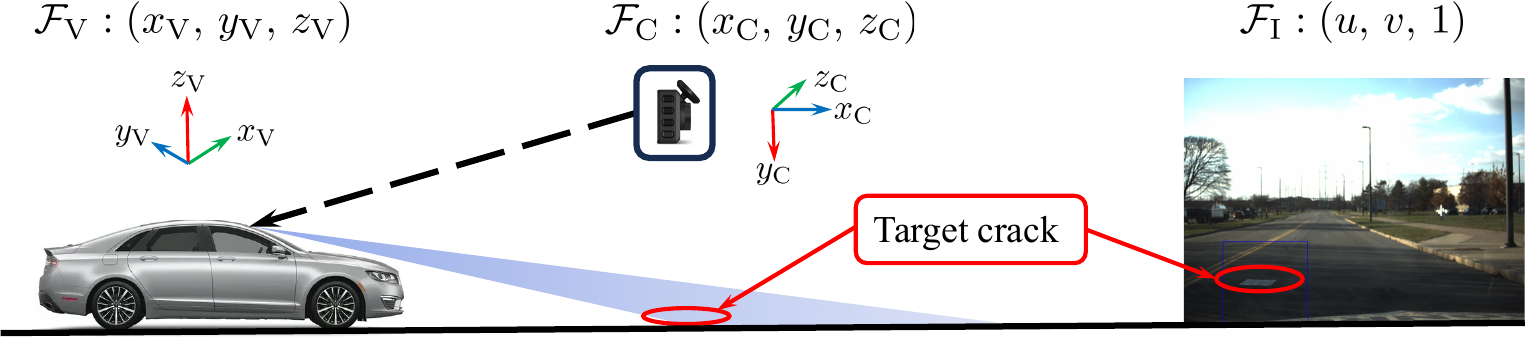}
    \caption{Vehicle frame to image frame transformation}
    \label{fig:coord_transformation}
\end{figure}
To determine cropping window location from the full image, the center of the crack is first projected from the 3D vehicle coordinate frame, \(\mathcal{F}_{\rm V} : (x_{\rm V},\, y_{\rm V},\, z_{\rm V})\), to the camera coordinate frame \(\mathcal{F}_{\rm C} : (x_{\rm C},\, y_{\rm C},\, z_{\rm C})\), and finally to the 2D image frame, $\mathcal{F}_{\rm I} : (u,\, v)$, using pinhole camera model. 
With the converted crack center pixel, the cropping window can be determined while remaining in the camera's field of view (FOV). 
In Fig.~\ref{fig:coord_transformation}, all these frames are illustrated.
Assumed that the crack lies on a flat road surface, the transformation from \(\mathcal{F}_{\rm V} : (x_{\rm V},\, y_{\rm V},\, z_{\rm V})\) to \(\mathcal{F}_{\rm C} : (x_{\rm C},\, y_{\rm C},\, z_{\rm C})\) is expressed as 
\begin{equation}
\label{eqn: V2C}
    \mathbf{p}_{\rm C} = R_{zyx}R_{\rm C}^{V}\, \mathbf{p}'_{\rm crack2C},
\end{equation}
where $R_{\rm C}^{V}$ denotes the fixed rotation from $\mathcal{F}_{\rm V}$ to $\mathcal{F}_{\rm C}$, and $R_{zyx} = R_z(\phi_{\rm C}) R_x(\theta_{\rm C}) R_y(\psi_{\rm C})$ represents the camera orientation defined by yaw $\phi_{\rm C}$, pitch $\theta_{\rm C}$, and roll $\psi_{\rm C}$.

Finally, to project the 3D crack center coordinate $\mathbf{p}_{\rm C}$ onto the image plane $\mathcal{F}_{\rm I}$, the intrinsic matrix $K$ of the camera is used :
\begin{equation}
\label{eqn:pinhole}
     \begin{bmatrix}
    u_{\rm C} \\
    v_{\rm C} \\
    1
    \end{bmatrix}
    =
    \frac{1}{z_{\rm C}}\, K\, \mathbf{p}_{\rm C}.
\end{equation}
An ROI of size $(B_{\rm w}, B_{\rm h})$ can be acquired with cropping corners
\begin{equation}\label{eqn:cropping_window}
\begin{aligned}
    u_{\min} &= \max\left(0, \; \min\left(u_{\rm C} - \frac{B_{\rm w}}{2},\;I_{\rm w}-B_{\rm w}\right)\right),\\
    v_{\min} &= \max\left(0, \; \min\left(v_{\rm C} - \frac{B_{\rm h}}{2},\;I_{\rm h}-B_{\rm h}\right)\right),\\
    u_{\max} &= u_{\min} + B_{\rm w}, \quad 
    v_{\max} = v_{\min} + B_{\rm h}.
\end{aligned}  
\end{equation}
We remark that the proposed design guarantees that for a perfectly calibrated camera, the true crack will always be inside the cropping window. 
While \eqref{eqn:cropping_window} gives a unique cropping window for a given frame, it is possible that for the same crack, multiple cropped images may be acquired. 
In Section \ref{subsec:pipeline_performance} we describe how the optimal cropping window with the best possible resolution may be selected; c.f., Stage III in Fig.~\ref{fig:design_overview}.

\subsection{Crack Length Estimation}
In this work, we seek to estimate crack length by computing the Euclidean distance between the two reconstructed endpoint points of the crack, expressed in the vehicle coordinate frame. 
The reconstruction is performed under the assumption that the crack lies on the flat road. 
We adopt this definition, consistent with the manual inspection instructions \cite{AASHTO2011bridgeInspection}, and to mitigate the zig-zag nature of fine cracks.

The accuracy of the length estimation depends highly on the camera extrinsic parameters, particularly the camera's mounting yaw $\phi_{\rm C}$, pitch $\theta_{\rm C}$, and roll $\psi_{\rm C}$, as these parameters directly determines the conversion between the vehicle and the camera coordinate frame. 
Thus, we first design a streamlined calibration of camera extrinsic parameters, specific to crack detection. Then we can proceed to proper length calculation.
 
\subsubsection{Camera Calibration}
\begin{figure}[t]
    \centering
    \includegraphics[width=\linewidth]{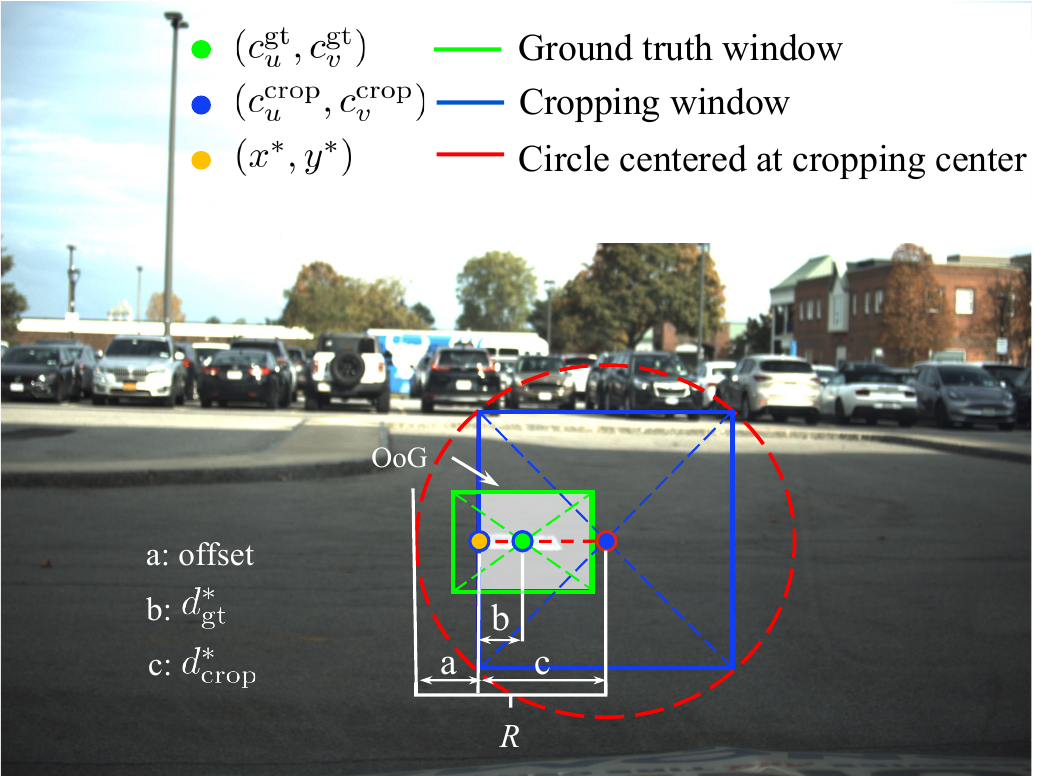}
    \caption{Camera calibration sample with key quantities used in the algorithm}
    \label{fig:image_calibration_schematics}
\end{figure}

Traditional camera calibration is done by formulating the calculation of intrinsic and extrinsic parameters as an optimization problem, where the objective function minimizes the reconstruction error between observed image points and their predicted projections computed from multiple viewpoints using the tunable camera parameters \cite{opencv_camera_calibration}. 
While the intrinsic parameters of the camera can be retrieved the manufacturer, the extrinsic parameters remain unknown. 
Calibrating vehicle camera extrinsic parameters could be challenging due to vehicle motions. 
In this work, we use a known calibration target with its center GPS position together with the known camera intrinsics. 
A checkerboard is chosen as the target due to its simplicity and reliability in pattern recognition practice. We further design the size of the cropping windows ($B_{\rm w}, B_{\rm h}$) to be big enough to cover the target object in a square box ($B_{\rm w} = B_{\rm h}$). 
Leveraging object detection model YOLO \cite{ultralytics_yolov8_docs}, accurate checkerboard bounding boxes can be obtained. 
We denote the center pixel of the ground truth window and the cropping window as $(c_u^{\rm gt}, c_{v}^{\rm gt})$ and $(c_u^{\rm crop},c_{v}^{\rm crop})$ respectively.
To align the cropped window with the ground truth window as close as possible, we use the unknown extrinsic as the optimization variable and an objective function with two metrics that can effectively quantify the alignment between the cropping windows
: overlapping with the ground truth ({\rm OoG}) and the alignment of traveling distance (AoT). 
The corresponding definitions are illustrated in Fig.~\ref{fig:image_calibration_schematics} and is detailed as follows.

{\bf Overlapping of Ground Truth (${\rm OoG}$)}:
 ${\rm OoG}$ measures ratio of ground truth $B^{\rm gt}$ has been included in the cropping window $B$.
\begin{equation}
    {\rm OoG}(c_u^{\rm crop},c_{v}^{\rm crop}) = \frac{\left| B \cap B^{\rm gt}\right|}{B^{\rm gt}}, \quad {\rm OoG}\in[0,1].
\end{equation}
    
{\bf Alignment of Traveling Distance (AoT)}: AoT measures the relative distance between two cropping windows.
For a square cropping window, the diagonal length can be interpreted as the diameter of a circle centered around the window center $(c_u^{\rm crop},c_{v}^{\rm crop})$, with radius $R$. 
The cropping window center corresponds to the theoretical projected pixel of the crack center. 
Since the ground true crack is always within the cropping box during calibration by design, one has $(c_x^{\rm gt}, c_y^{\rm gt}) \in
    \bigl[c_x^{{\rm crop}}-\tfrac{L_{\rm crop}}{2},\, c_x^{{\rm crop}}+\tfrac{L_{\rm crop}}{2}\bigr]$ . 
The circle region can then be naturally used as a reference for evaluating the spatial relation between $(c_u^{\rm crop},c_{v}^{\rm crop})$ and $(c_u^{\rm gt}, c_{v}^{\rm gt})$. 
A pixel, $(x^*,y^*)$, on the edge of the cropping window that lies on the line connecting the two center pixels and shares the same slope can be computed as:
\begin{equation}
\begin{aligned}
d_{cx} &= \mathbf{max}(c_u^{\rm gt} - c_u^{\rm crop}, \epsilon), \\
d_{cy} &= \mathbf{max}(c_{v}^{\rm gt} - c_{v}^{\rm crop}, \epsilon), \\
s &= \mathbf{min}\left(\frac{\frac{L_{\rm crop}}{2}}{|d_{cx}|},\; \frac{\frac{L_{\rm crop}}{2}}{|d_{cy}|}\right),\\
(x^*,y^*) &= (c_u^{\rm crop},c_{v}^{\rm crop}) + s\,(d_{cx},d_{cy}).
\end{aligned}
\end{equation}
The corresponding distances between the projected edge pixel and the cropping window center, as well as between the projected edge pixel and the ground-truth center, are computed as
\begin{equation}
\begin{aligned}
    d^{*}_{\rm crop} &= {\sqrt{\left({\left(x^*\right)}^2 - {\left(c_u^{\rm crop}\right)}^2\right) 
    + \left({\left(y^*\right)}^2 - {\left(c_{v}^{\rm crop}\right)}^2\right)}},\\
    d^{*}_{\rm gt} &= {\sqrt{\left({\left(x^*\right)}^2 - {\left(c_u^{\rm gt}\right)}^2\right) 
    + \left({\left(y^*\right)}^2 - {\left(c_{v}^{\rm gt}\right)}^2\right)}}.
\end{aligned}
\end{equation}
The offset between the cropping window boundary and the circle shared the same center is then defined as: 
\begin{equation}
    {\rm offset} = \frac{L_{\rm crop}}{2} - d^{*}_{\rm crop}.
\end{equation}
The Alignment of Traveling Distance is finally defined as
\begin{equation}
    {\rm AoT} = \frac{d^{*}_{\rm gt} + {\rm offset}}{R}
\end{equation}
A higher ${\rm AoT}$ indicates that the ground-truth center $(c_u^{\rm gt}, c_{v}^{\rm gt})$ is closer to the cropping window center $(c_u^{\rm crop},c_{v}^{\rm crop})$. 

With these two metrics, the full optimization can be set up as: 
\begin{equation}
\label{eqn:cropping_performance}
\begin{aligned}
\max_{\phi_{\rm C}, \theta_{\rm C}, \psi_{\rm C}} \quad
& {{\rm OoG}}\!\left(c_u^{{\rm crop}}, c_{v}^{{\rm crop}}\right)
+ {\rm AoT}\!\left(c_u^{{\rm crop}}, c_{v}^{{\rm crop}}\right) \\
\text{s.t.} \quad
& \left| c_u^{\rm gt} - c_u^{{\rm crop}} \right|
\le \frac{L_{{\rm crop}}}{2}, \\
& \left| c_{v}^{\rm gt} - c_{v}^{{\rm crop}} \right|
\le \frac{L_{{\rm crop}}}{2}, \\
& 0 \le {{\rm OoG}}\!\left(c_u^{{\rm crop}}, c_{v}^{{\rm crop}}\right) \le 1 .
\end{aligned}
\end{equation}
In this work we approximate the solution to \eqref{eqn:cropping_performance} using the bisection method with a resolution of 1 degree.

\subsubsection{Real-time length calculation}\label{sec:length_estimation}
To accurately estimate the length between the crack endpoints, an edge-corner detection algorithm is developed. 
The algorithm takes the crack mask image as input and details on how the crack mask is generated is detailed in Section \ref{sec:crack_detection_model}.
It extracts pixels that are put crack mask and stays within the crack region. 
Each extracted pixel is then reconstructed from the image coordinate frame to the vehicle coordinate frame. 
This reconstruction follows the reverse calculation from \eqref{eqn:pinhole}, with intrinsic matrix $K$ given by
\begin{equation}
\begin{bmatrix}
f_x & 0 & c_x \\
0 & f_y & c_y \\
0 & 0 & 1
\end{bmatrix},
\end{equation} 
where $f_x \text{ and } f_y$ denote the focal length along the camera axes $x_{\rm C} \text{ and } y_{\rm C}$, $c_x \text{ and } c_y$ denote the central pixel in $\mathcal{F}_{\rm I}$.
We first project the 2D image coordinate back to a normalized 3D camera frame \(\mathcal{F}_{\rm C} : (x_{\rm C},\, y_{\rm C},\, z_{\rm C})\), 
\begin{equation}
    \mathbf{C}_{\text{norm}, \mathcal{F}_{\rm C}} = \begin{bmatrix} \dfrac{u - c_x}{f_x} & \dfrac{v - c_y}{f_y} & 1 \end{bmatrix}^{\rm T}.
\end{equation}
The normalized coordinates in the camera frame are then transformed into the vehicle frame $\mathcal{F}_{\rm V} : (x_{\rm V},\, y_{\rm V},\, z_{\rm V})$ using: 
\begin{equation}
    \mathbf{C}_{\text{norm}, \mathcal{F}_{\rm V}} = (R_{\rm C}^{\rm V})^{\rm T} \, R_{zyx}^{\rm T} \, \mathbf{C}_{\text{norm}, \mathcal{F}_{\rm C}}
\end{equation}
We remark that the reconstructed pixel is normalized so it must be scaled back to be accurately expressed in the vehicle frame. 
Recall the camera offset and the assumption that the cracks lie on the flat ground surface, the de-normalized coordinates are obtained using:
\begin{align}
    \mathbf{C}_{\mathcal{F}_{\rm V}} &= \mathbf{t}_{\rm VC} + C_{z} \, \mathbf{C}_{\text{norm}, \mathcal{F}_{\rm V}} \\
    C_{z} &= -\frac{(\mathbf{t}_{\rm VC})_z}{(\mathbf{C}_{\text{norm}, \mathcal{F}_{\rm V}})_z},
\end{align}
where $\mathbf{t}_{\rm VC}$ denotes the (assumed known) offset of the camera lens with respect to the RTK GPS antenna, expressed in the vehicle frame. 
The reconstructed points are subsequently divided into four quadrants with respect to the crack center, and edge locations are selected as the pixels with the maximum Euclidean distance from the center within each quadrant.
The edge selection algorithm is summarized in Algorithm \ref{algo:Edge_Selection}.

\begin{algorithm}[H]
\caption{Edge Corner Selection}\label{algo:Edge_Selection}
\begin{algorithmic}[1]
\Require Highlighted pixel set $\{x_k\}_{k=1}^N$
\Ensure Quadrant correlation vector ${\rm Corr}$, edge coordinates $\{E_q\}$

\State Extract $N$ highlighted pixels
\For{$k = 1$ to $N$}
    \State Transform $x_k$ to vehicle-frame coordinates
\EndFor

\State Compute center coordinate $x_{\rm C}$ of $\{x_k\}$
\State Initialize dictionary $E \gets \emptyset$
\State Initialize vector ${\rm Corr} \in \mathbb{R}^4$

\For{$q = 1$ to $4$}
    \State $\mathcal{X}_q \gets \{x_k \mid x_k \in \text{quadrant } q\}$
    \If{$\mathcal{X}_q \neq \emptyset$}
        \State $x_k^\ast \gets \arg\max_{x_k \in \mathcal{X}_q} \|x_k - x_{\rm C}\|_2$
        \State $E_q \gets \{x_k^\ast\}$
        \State ${\rm Corr}_q \gets 1$
    \Else
        \State ${\rm Corr}_q \gets 0$
    \EndIf
\EndFor
\end{algorithmic}
\end{algorithm}


\section{Crack Detection Model}
\label{sec:crack_detection_model}
In this section, we describe the adaptation of the existing crack detection models for real-time images from vehicle's front view camera.
In this work, we focus on improving the generalization under real-world operational condition. 
We considered two leading open-source semantic segmentation architectures for crack detection given their balanced accuracy and real-time performance \cite{ma2025vehicular}: DeepCrack, a multi-scale convolution encoder-decoder model \cite{zou2018deepcrack}, is employed using its pre-trained weights without modifications; LECSFormer, a transformer-based segmentation based model \cite{chen2022refined}, is adapted and improved for our real-time detection pipeline. 

\subsection{Model Reconstruction}
To fill missing details and recover baseline model weight for standard dataset, we reconstruct and further refine the LECSFormer training framework, with focuses on loss function and training data augmentation. 
\subsubsection{Loss function}
One major difficulty in crack segmentation is the class imbalance, where crack pixels only take a small fraction of the image. 
To address this, the original LESCFormer uses Binary Cross-Entropy (BCE) loss with high penalty weight to the crack pixels using $\omega_{\text{pos}} = 5.0$ and $\omega_{\text{neg}} = 1.0$:

\begin{equation}
\mathcal{L}_{\text{wBCE}} = \frac{1}{N} \sum_{i=1}^{N} \omega_i \Big[ 
- y_i \log \sigma(\hat{y}_i) 
- (1 - y_i)\log \big(1 - \sigma(\hat{y}_i)\big) 
\Big],
\end{equation}
where $\hat{y}_i$ denotes the predicted logits, $y_i \in \{0,1\}$ is the ground truth, $\sigma(\cdot)$ is the sigmoid function, and $\omega_i$ is selected based on class label.
To further optimize region-level overlap, we includes the Dice loss:
\begin{equation}
\mathcal{L}_{\text{Dice}} = 1 - 
\frac{2 \sum_{i=1}^{N} \sigma(\hat{y}_i) y_i + \epsilon}
{\sum_{i=1}^{N} \sigma(\hat{y}_i) + \sum_{i=1}^{N} y_i + \epsilon}.
\end{equation}
We use $\epsilon = 1.0$ for numerical stability.
The final loss is a weighted combination of both terms:
\begin{equation}
\mathcal{L}_{\text{combined}} =
\lambda_{\text{BCE}} \mathcal{L}_{\text{wBCE}} +
\lambda_{\text{Dice}} \mathcal{L}_{\text{Dice}}
\end{equation}
where $\lambda_{\text{BCE}} = 0.7$ and $\lambda_{\text{Dice}} = 0.3$ are used.

Following the original multi-scale supervision strategy in \cite{chen2022refined}, this combined loss is applied to the final output and all intermediate decoder outputs:
\begin{equation}
\mathcal{L}_{\text{total}} =
\mathcal{L}_{\text{combined}}^{\text{out}} +
\sum_{k=1}^{K} \mathcal{L}_{\text{combined}}^{\text{mid},k},
\end{equation}
where $K$ denotes the number of supervised decoder stages. This formulation improves robustness to class imbalance while preserving multi-scale feature consistency.

\subsubsection{Image augmentation}
\label{sec: image aug}
Transformer based models such as LECSFormer require large and diverse training data to generalize effectively, but the existing crack datasets are suffered from the size and inconsistent visibility. 
To address this, we design a crack-density-aware augmentation strategy, in which training samples are categorized into four groups (none, minimal, moderate, dense) based on the number of crack appearances in each sample image. 
Categories with fewer samples are considered as underrepresented, and will be augmented using geometric transformation, photometric variation, elastic deformation and density-aware cropping to upscale their representation by a factor of 2. 

\subsubsection{Reconstruction results}
To evaluate our reconstructed LECSFormer model training pipeline performance, we retrain the model on the augmented CrackTree260 dataset and further benchmark performance on the CRKWH100 dataset (used in the literature to check model generalization capabilities \cite{ma2025vehicular}).
The reproduced benchmark results, together with the reported
metrics from the original work, are summarized in Table~\ref{tab:crkwh100_comparison}. 
Overall, we are able to reproduce similar performance, despite using different augmentation methods.

\begin{table}[t]
\centering
\caption{Benchmark Performance on CRKWH100 Dataset}
\label{tab:crkwh100_comparison}
\setlength{\tabcolsep}{4pt}
\renewcommand{\arraystretch}{1.1}
\begin{tabular}{|l|c|c|c|}
\hline
\textbf{Model} &
\textbf{Train Data} &
\textbf{ODS F1} &
\textbf{AP} \\ \hline

LECSFormer~\cite{chen2022refined} &
CT260+CLS315 &
0.9530 &
Not-provided \\ \hline



LECSFormer (Ours) &
CT260 &
0.9184 &
0.4973 \\ \hline



\end{tabular}
\end{table}

\begin{table}[t]
\centering
\caption{Performance on Real-Time Front View Camera Dataset (RT100)}
\label{tab:realtime_performance}
\begin{tabular}{|l|c|c|c|c|}
\hline
\textbf{Model} &
\textbf{ODS F1} &
\textbf{Precision} &
\textbf{Recall} &
\textbf{AP} \\ \hline

LECSFormer &
0.3744 &
0.3425 &
0.4605 &
0.1371 \\ \hline

DeepCrack &
0.3374 &
0.2912 &
0.4359 &
0.1225 \\ \hline

\end{tabular}
\end{table}

\subsection{Front View Camera Image Performance}
To evaluate real-world performance from the vehicle's front view camera images, we collect a dataset of 443 images recorded under diverse operational conditions, including different lighting, glare, motion blur, and asphalt textures. The dataset is split into 343 training images (denoted as RT 343) and 100 testing images (denoted as RT 100). 
We adopt a hybrid annotation pipeline to accelerate the training set annotation: in addition to manual annotation, we used black hat filter \cite{opencv_camera_calibration} to generate initial crack mask, and manually refine the mask afterwards. 

As shown in Table~\ref{tab:realtime_performance}, although our trained models have shown strong benchmark performance, both models has shown degraded performance on RT 100. 
This performance degradation is not surprising considering that the viewing angle is significantly different, but it highlights the need on improving the model generalization capability to different viewing angles. 
To address this, we incorporate RT343 into the training process to adapt the models to vehicle's front view camera. 

\begin{figure}[t]
\centering
\includegraphics[width=\linewidth]{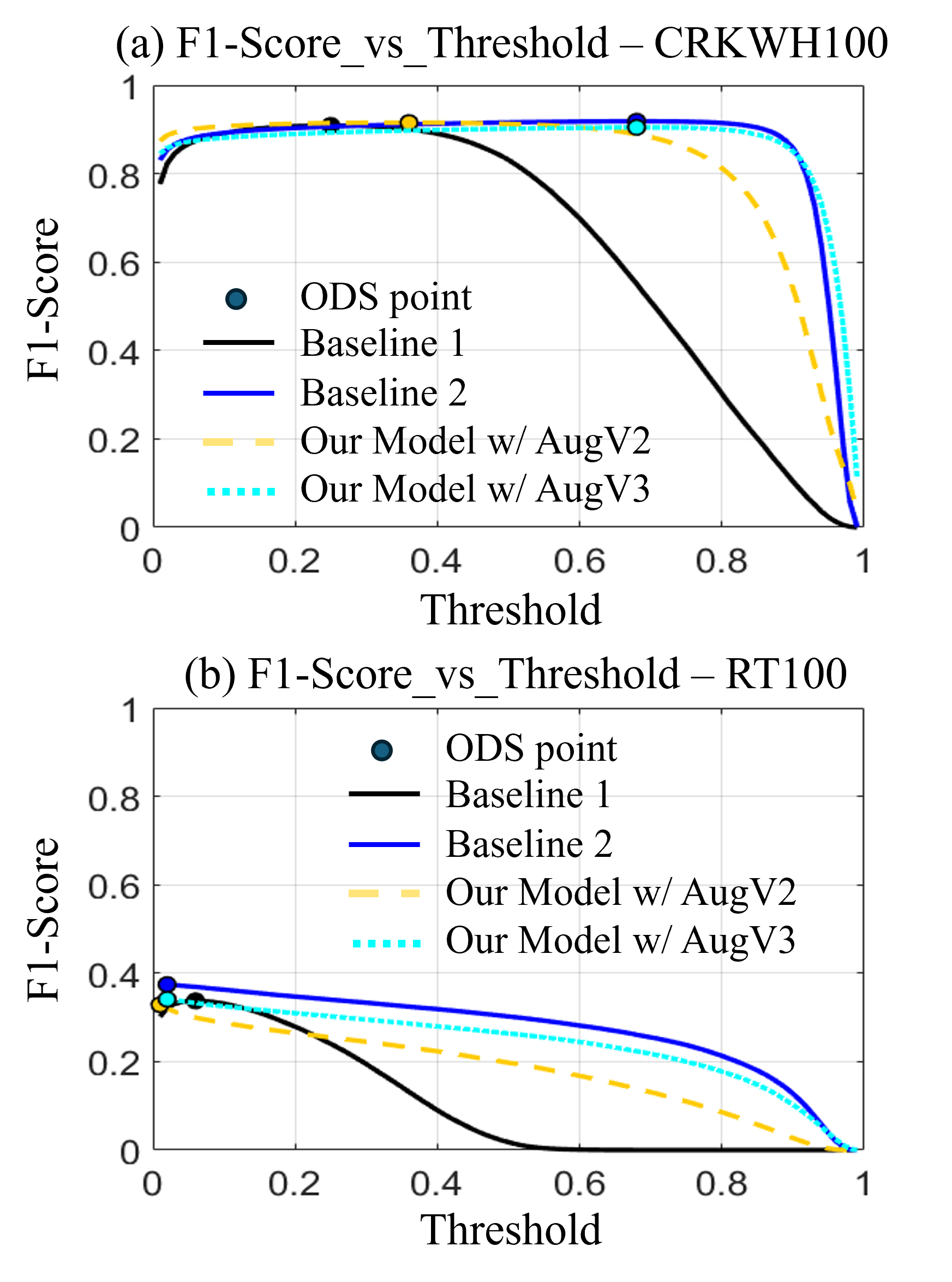}
\caption{F1-threshold curves across datasets: (a) CRKWH100 benchmark and (b) RT100 test set comparing five configurations: Baseline 1 (DeepCrack pre-trained), Baseline 2 (LECSFormer trained on CT260 with advanced augmentation), Our Model AugV2 (CT260 with DeepCrack augmentation), and Our Model AugV3 (CT260 with enhanced DeepCrack augmentation). The dots indicate ODS values, and Baseline 2 shows the best robustness across threshold variations on real-time data.}
\label{fig:f1_threshold_curves}
\end{figure}

\begin{figure}[t]
\centering
\includegraphics[width=\linewidth]{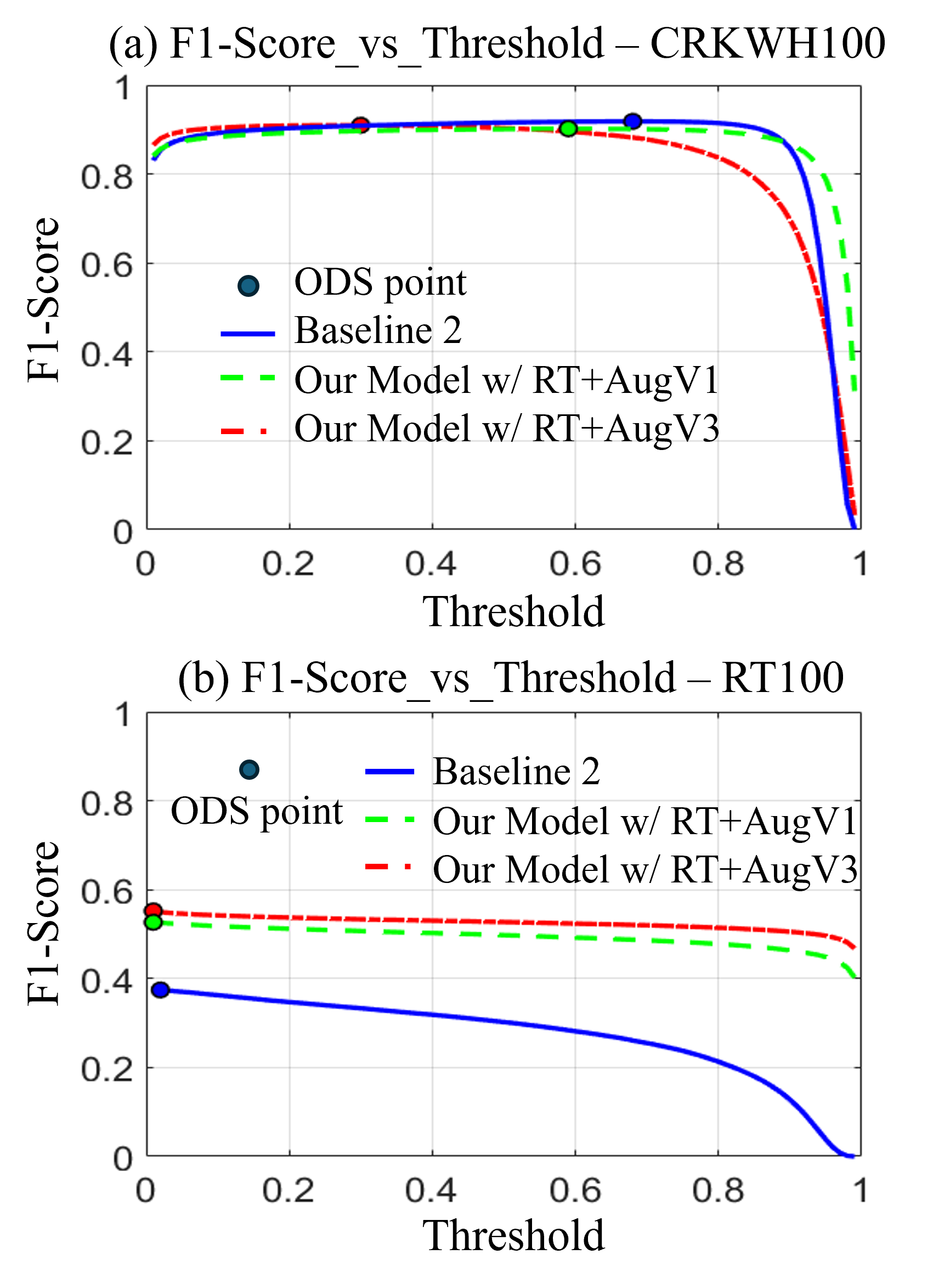}
\caption{F1-threshold curves on the (a) CRKWH100 dataset and (b) RT100 test set for models incorporating front view camera image: Baseline 2 (LECSFormer trained on CT260 with advanced augmentation), Our Model w/ RT+AugV1 (trained on CrackTree260 + real-time images with Advanced Augmentation V1), and Our Model w/ RT+AugV3 (trained on CrackTree260 + real-time images with Enhanced DeepCrack Augmentation V3). Integration of domain-specific data with sophisticated augmentation yields significantly higher and more stable performance across thresholds.}
\label{fig:f1_threshold_curves_rt}
\end{figure}

\subsection{Ablation Study}
To evaluate the impact of image augmentation and front view camera image incorporation, we conduct a systematic ablation study in which the augmentation strategy and training dataset are varying factors. 
The model performance is evaluated on both the RT100 and CRKWH100 testing sets, and the results are concluded in Table~\ref{tab:ablation_results}. 

\begin{table}
\centering
\caption{Ablation Study: Real-Time Performance (ODS F1)}
\label{tab:ablation_results}
\small
\begin{tabular}{|p{5.8cm}|c|}
\hline
\textbf{Training Configuration} & \textbf{ODS (RT100)} \\
\hline
\multicolumn{2}{|c|}{\textbf{Baseline Models}} \\
\hline
\quad DeepCrack (pre-trained) ({\bf Baseline 1}) & 0.3374 \\
\quad LECSFormer (CT260 - no augmentation) & 0.2923 \\
\hline
\multicolumn{2}{|c|}{\textbf{Augmentation Effect (CrackTree260)}} \\
\hline
\quad DeepCrack augmentation (V2) & 0.3287 \\
\quad Enhanced DeepCrack augmentation (V3) & 0.3411 \\
\quad Advanced augmentation (V1) ({\bf Baseline 2}) & 0.3744 \\
\hline
\multicolumn{2}{|c|}{\textbf{Front View Camera Image Integration}} \\
\hline
\quad CT260 + RT343, Aug V1 ({\bf Our model 1})& 0.5267 \\
\quad CT260 + RT343, Aug V3 ({\bf Our model 2}) & \textbf{0.5522} \\
\hline
\end{tabular}
\end{table}

We begin with baseline models trained without augmentation. Specifically, DeepCrack with pre-trained weights is used as Baseline 1, and our reconstructed LECSFormer trained on CrackTree260 without augmentation establishes a performance floor of 0.2923 ODS on RT100.

We then evaluate the effect of augmentation by applying three strategies on the CrackTree260 dataset. 
Our augmentation strategy introduced in Section~\ref{sec: image aug} is denoted as V1, the augmentation strategy originally described in DeepCrack is denoted as V2, an enhanced version of V2 with additional transformation is denoted as V3, respectively. 
While all augmentation methods improve performance over the no-augmentation cases, V1 and V3 outperform V2 on RT100 with higher ODS scores. 
Since V3 subsumes the transformations used in V2, we exclude V2 from further analysis. 

In the subsequent study, we focus on V1 and V3 for evaluating the impact of front view camera image integration. 
The model trained on CrackTree260 with V3 augmentation is used as Baseline 2 for comparison. 
As shown in Table~\ref{tab:ablation_results}, adding the front view camera image into the training set further improved the model performance for both the augmentation strategy V1 and V3. 
The most significant improvement is achieved when combining front view camera images with the enhanced V3 augmentation, achieving a peak ODS of 0.5522. This is an 89\% improvement over the non-augmented LECSFromer trained on CrackTree260 dataset.

In Figs.~\ref{fig:f1_threshold_curves} and~\ref{fig:f1_threshold_curves_rt}, we present the F1-score curves of all models listed in Table~\ref{tab:ablation_results} to demonstrate their performance on both the benchmark testing set (CRKWH100) and the real-time testing set(RT100).
Fig.~\ref{fig:f1_threshold_curves} compares the performance between Baseline 1 (DeepCrack with pre-trained weights) and LESCFormer models trained using three augmentation strategies. 
It shows that models trained with higher augmentation factors (Baseline 2 and V3) outperform the model trained  with less augmentation factor (V1) on both two testing sets. \\
Fig.~\ref{fig:f1_threshold_curves_rt}, on the other hand, further compares the model performance between the models trained with real-time dataset incorporation (Our Model w/ RT+ AugV1 and Our Model w/ RT+AugV3) and Baseline 2. 
It shows a mixed trend: Incorporating real-time training set slightly degrades the  model performance below the Baseline 2 on the benchmark testing set (CRKWH100), but significantly improved the performance on RT100. 
Specifically, F1-score improves from (0.2-0.4) to (0.4-0.6) within threshold range of (0-0.8). 
This finding underscores the importance of domain-specific data is in closing the generalization gap to different view images.

\subsection{Model Selection}
Our study on adapting crack detection models to vehicle front view camera images indicates that the benchmark performance does not directly translate to the real-world deployment. 
As shown in Table~\ref{tab:crkwh100_comparison} and Table~\ref{tab:realtime_performance}, models achieving high accuracy on curated dataset show a significant performance degradation when applied to front view camera images. 
Furthermore, while image augmentation improves the model robustness, it is not sufficient alone. 
Although all augmentation strategies enhance performance, their impact remains limited without adding domain-specific data. 
Performance are improved combining both the augmentation strategy with domain-specific real-world training data; thus, we select the model trained with augmentation V1 (our model 1) and V3 (our model 2) as our final models to be used in the full crack detection pipeline.


\section{Full pipeline implementation}\label{sec:pipeline}
Building on the CV2X view selection algorithms and crack detection model reported in the previous sections, we are ready to establish a complete infrastructure-guided connectivity-enhanced road crack detection and estimation pipeline. 
In this section, we first explain the workflow of our full pipeline shown in Fig.~\ref{fig:design_overview}. 
To validate our detection results, we used a widely used ground-truth measurement approach with a depth camera, ZED2, and compared our crack length estimates with manual inspection results to present our findings. 

\subsection{Full pipeline overview}
The complete detection pipeline requires close cooperation between the C-V2X road side unit (RSU) and the software stack that includes C-V2X on-board unit (OBU) on our CAV platform. 
The software architecture is developed based on ROS2\cite{ros2_foxy} and Autoware \cite{autoware_2023}. 
It contains four carefully designed ROS2 nodes that support real-time detection: the OBU node, the Vehicle node, the Camera node, and the Detection node. 
Upon receiving a detection request from the RSU, the four nodes collaborate through three sequential stages to complete the detection task and send the result back to the RSU. 

\subsubsection{Stage $\mathbf{I}$: Pre-detection}
\label{sssec: pred}
\begin{figure}[t]
    \centering
    \includegraphics[width=0.35\textwidth]{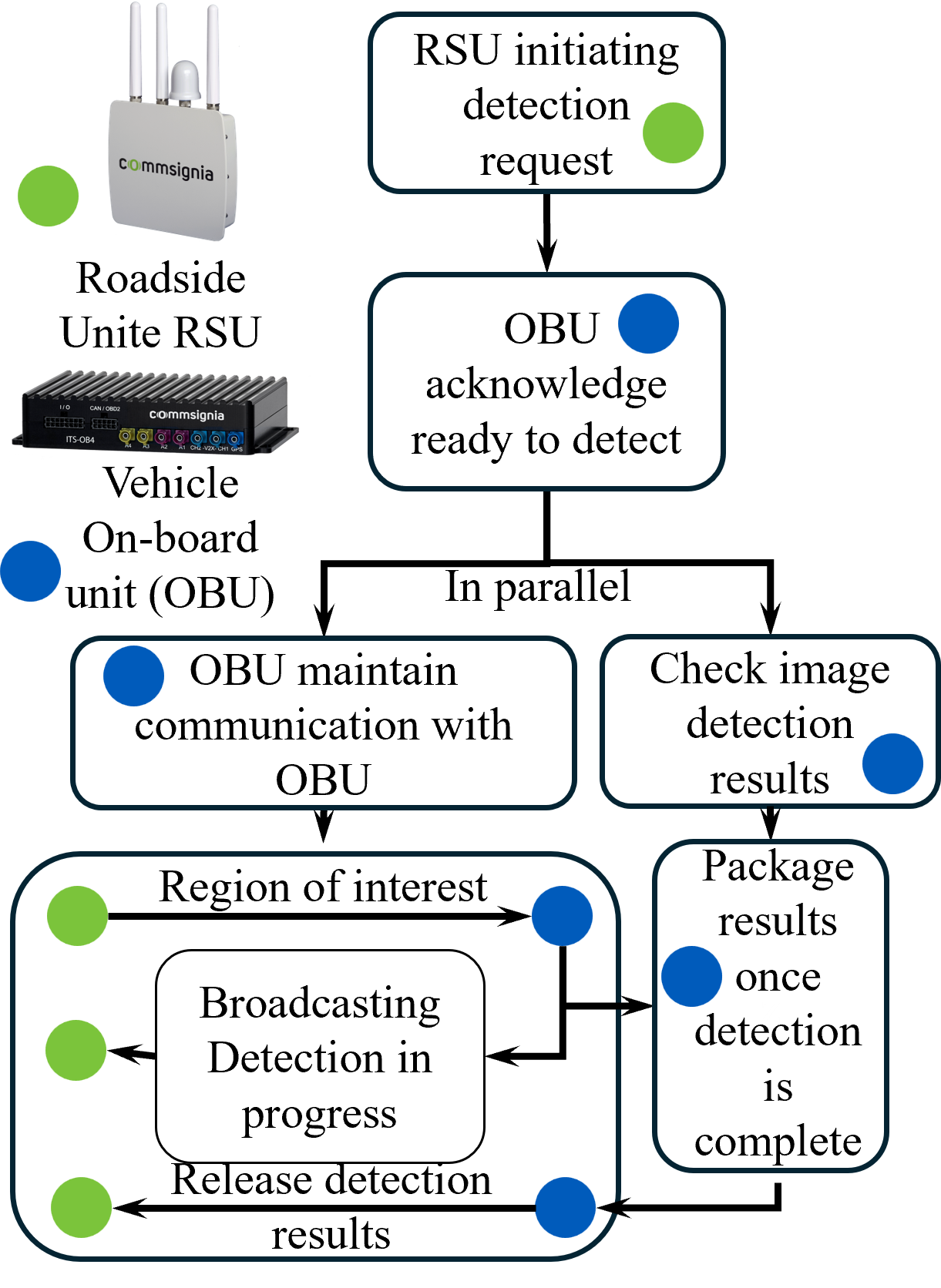}
    \caption{RSU--OBU communication protocol used for AOI dissemination and feedback exchange with a vehicle-resident crack detection pipeline.}
    \label{fig:protocol_design}
\end{figure}

Stage $\mathbf{I}$ assigns detection tasks to the CAV through a specially designed RSU-initiated request–response workflow protocol between the RSU and the OBU, as shown in Fig.~\ref{fig:protocol_design}. 
The RSU is continuously looking for the signals from the OBU running on the CAV. 
Once the CAV enters the RSU communication range and gets detected, the RSU transmits the crack center coordinates along with relevant metadata. Meanwhile, the OBU node starts sending the received crack coordinate information along with the traffic data, including vehicle information from surrounding CAVs and the status of the ego CAV itself, to the vehicle node. 
This step enables CAV to navigate to the crack location for detection and send back the result once the detection is done. 
\subsubsection{Stage $\mathbf{II}$: In-detection}
\label{sssec: ind}
In parallel with subscribing to the crack location, the vehicle node also subscribes to all the CAV statuses from both the OBU node and all the onboard sensor flows. 
With real-time CAV location along with the crack location, the vehicle node computes the distance between the crack and the front-view camera lens using \eqref{eqn: V2C}. 
This distance is further used to decide whether the crack resolution is above the threshold for further detection model inference. 
\begin{table*}[t]
\centering
\caption{Comparison of crack length estimation under different resolutions.
Ground truth length = 3.24~m; ZED2 reference = 3.09~m.}
\label{tab:resolution_comparison}
\renewcommand{\arraystretch}{1.15}
\setlength{\tabcolsep}{4pt}

\begin{tabular}{
>{\centering\arraybackslash}m{2.3cm}
>{\centering\arraybackslash}m{2.4cm}
>{\centering\arraybackslash}m{2.4cm}
>{\centering\arraybackslash}m{2.4cm}
>{\centering\arraybackslash}m{2.4cm}
>{\centering\arraybackslash}m{3.2cm}
}
\hline
\textbf{Model} &
\textbf{Target Crop} &
\textbf{Mask (Low)} &
\textbf{Mask (Mid)} &
\textbf{Mask (High)} &
\textbf{Estimated Length [m]} \\

  &
  &
  &
  &
  &
\textbf{(Low / Mid / High)} \\
\hline

Baseline 1
& \includegraphics[width=2.4cm]{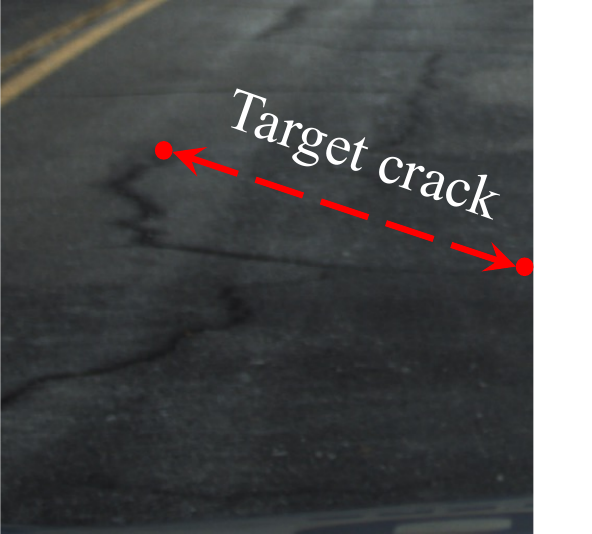}
& \includegraphics[width=2.1cm]{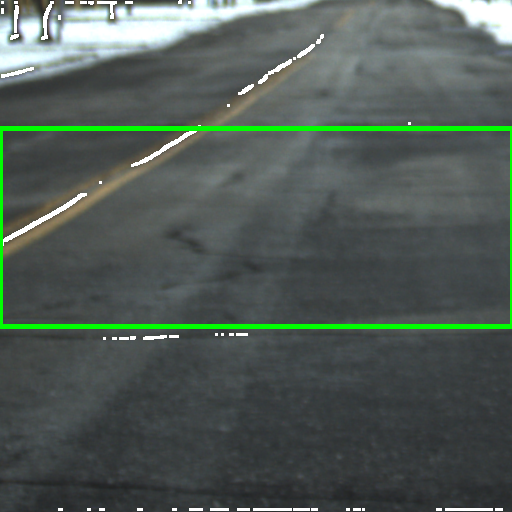}
& \includegraphics[width=2.1cm]{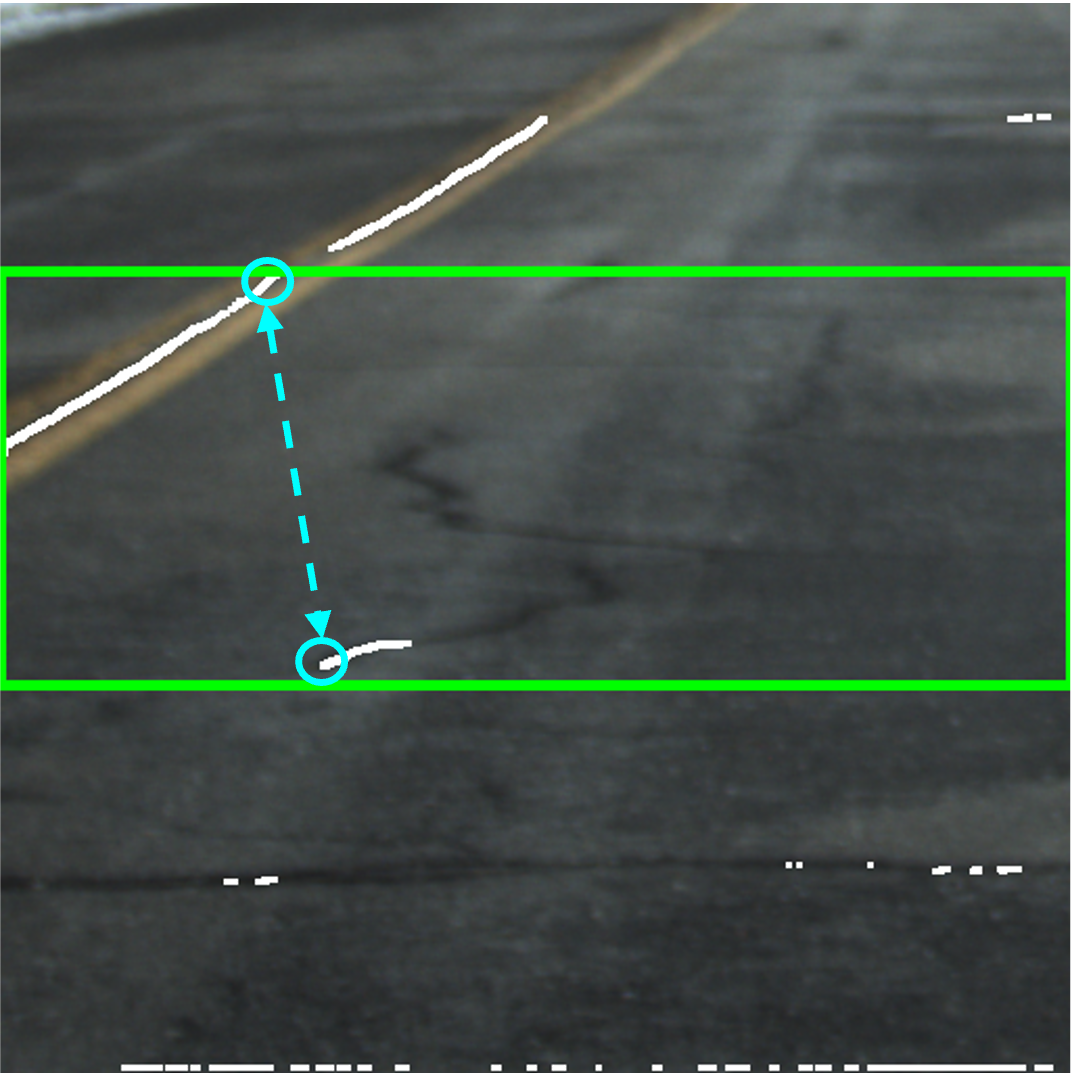}
& \includegraphics[width=2.1cm]{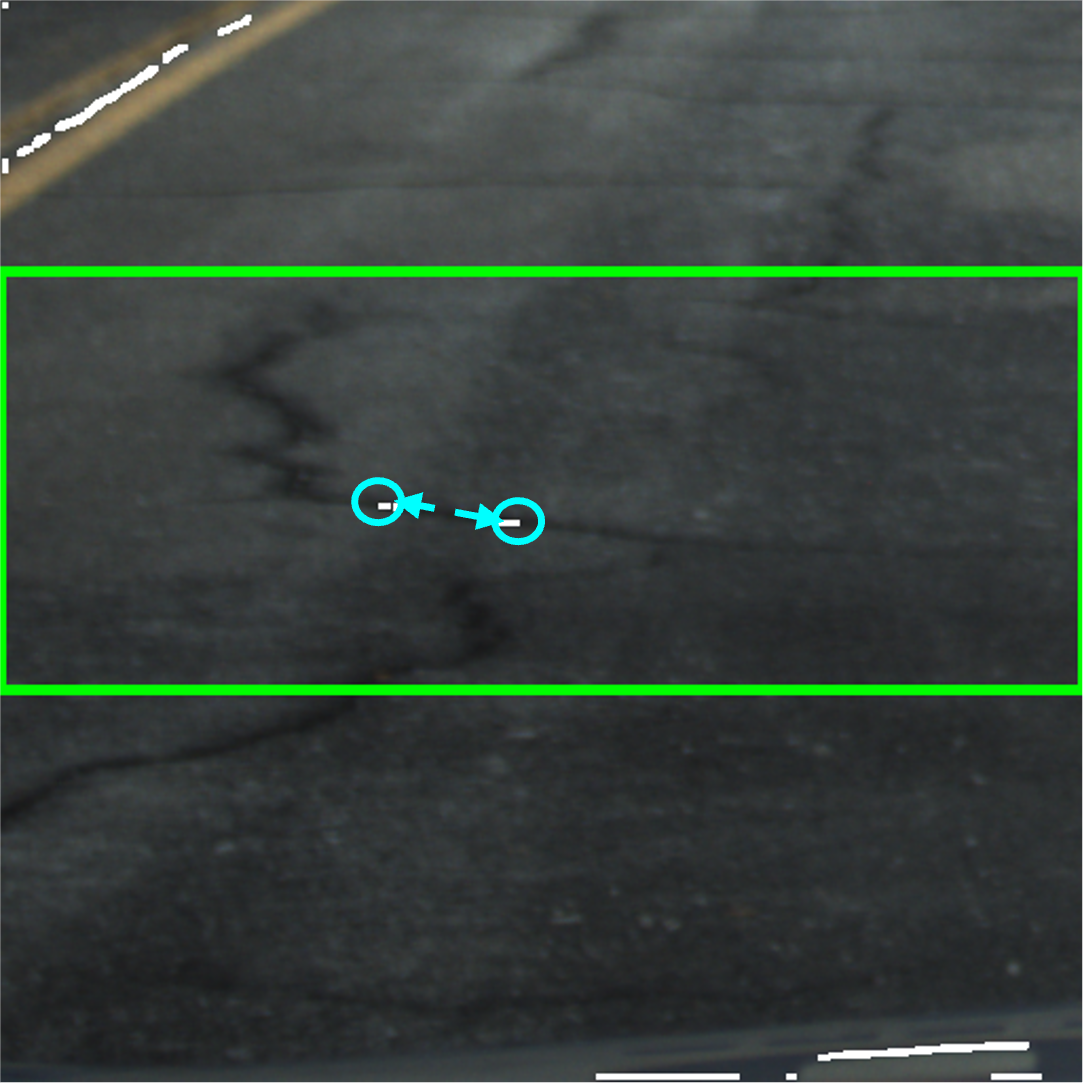}
& NA / 8.0069 / 0.2464 \\

Baseline 2
& \includegraphics[width=2.4cm]{Figures/Target_crack.pdf}
& \includegraphics[width=2.1cm]{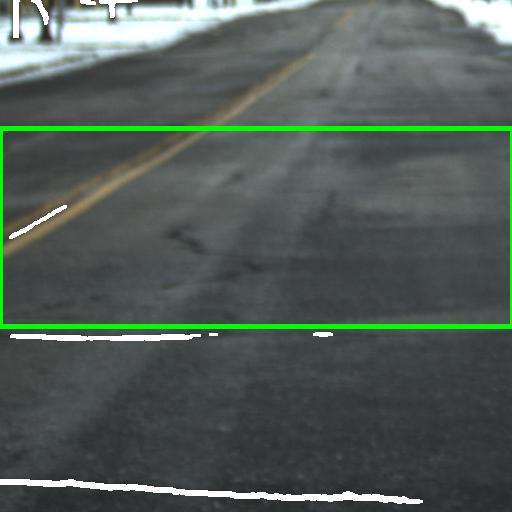}
& \includegraphics[width=2.1cm]{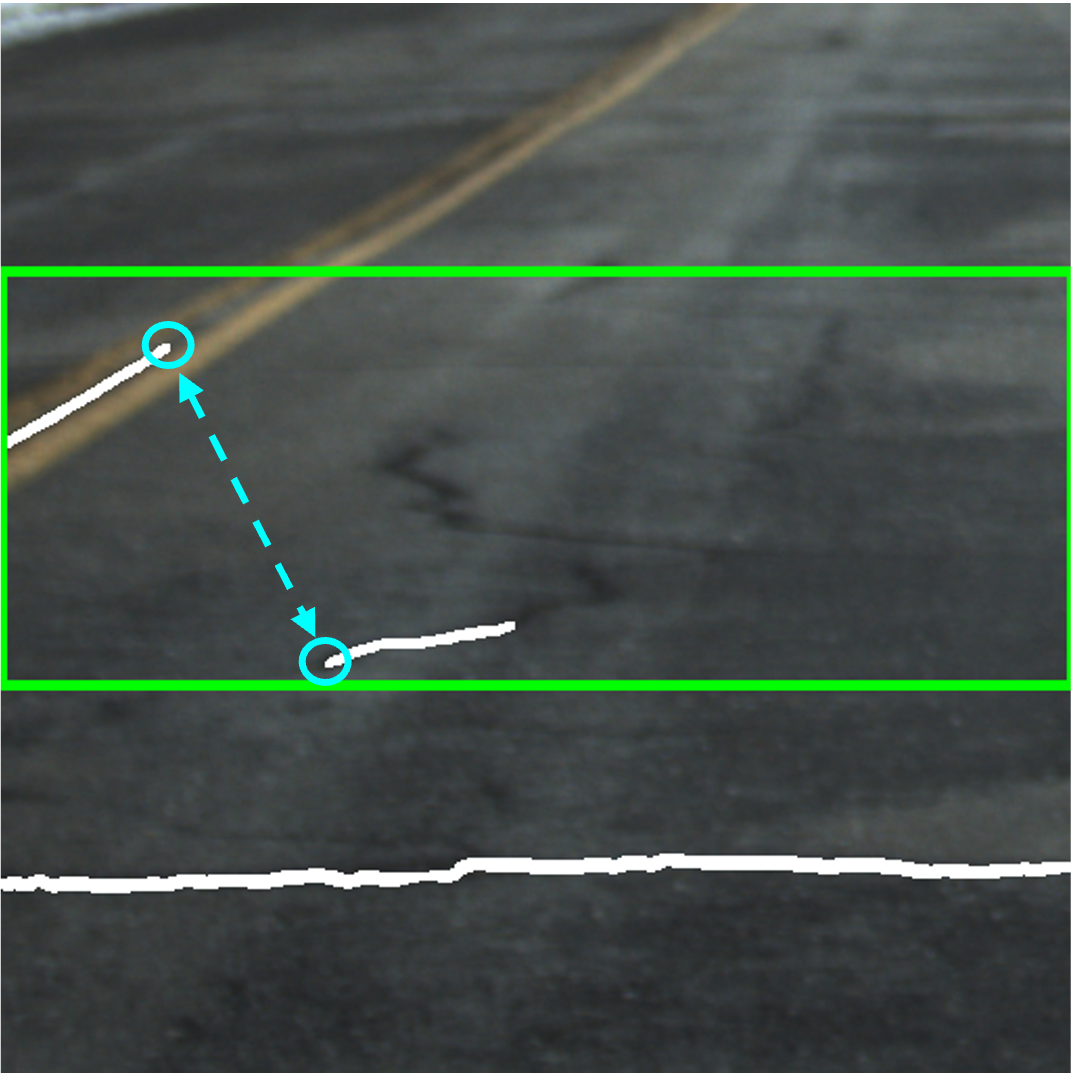}
& \includegraphics[width=2.1cm]{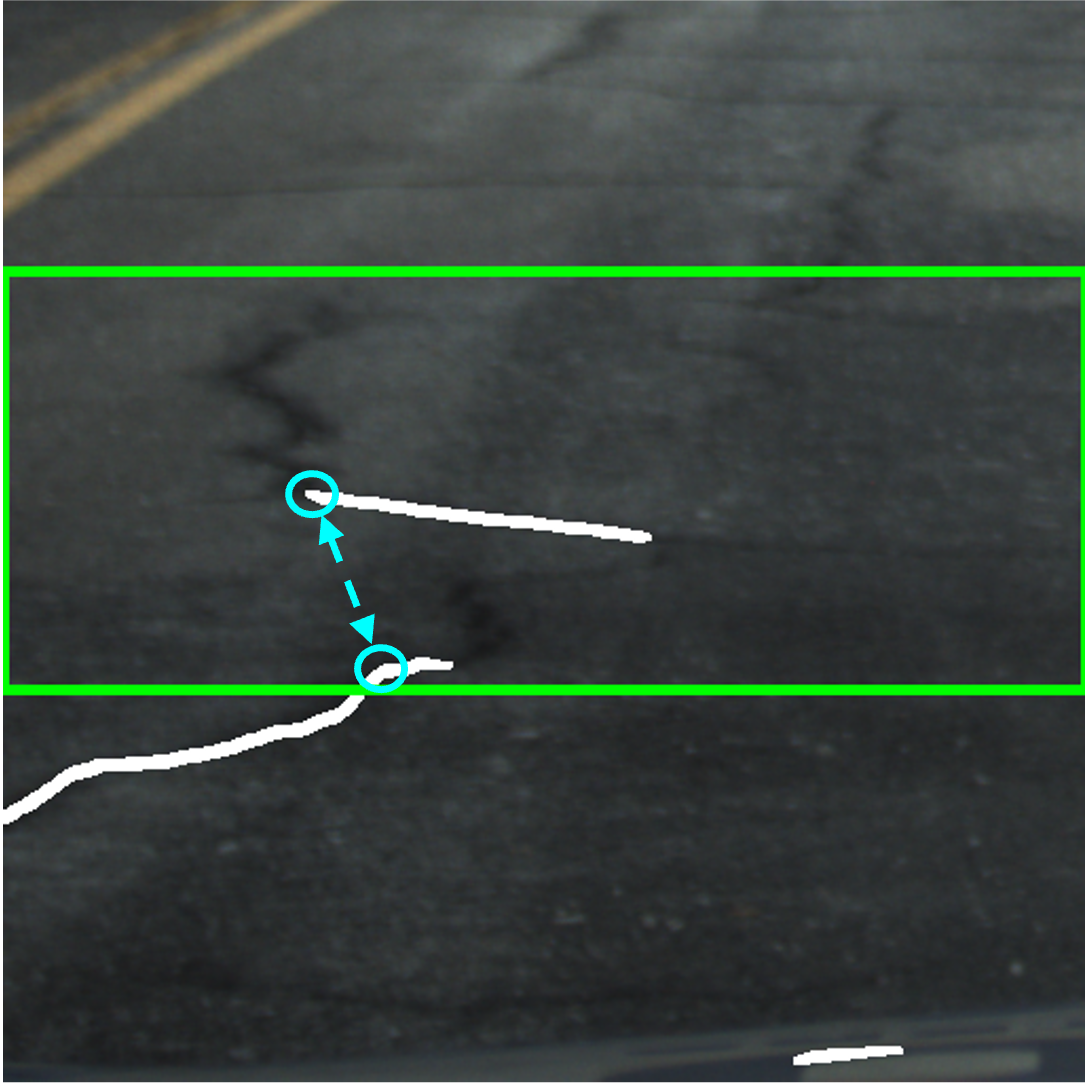}
& NA / 5.7899 / 1.5224 \\

Our Model w/ RT + Aug V1
& \includegraphics[width=2.4cm]{Figures/Target_crack.pdf}
& \includegraphics[width=2.1cm]{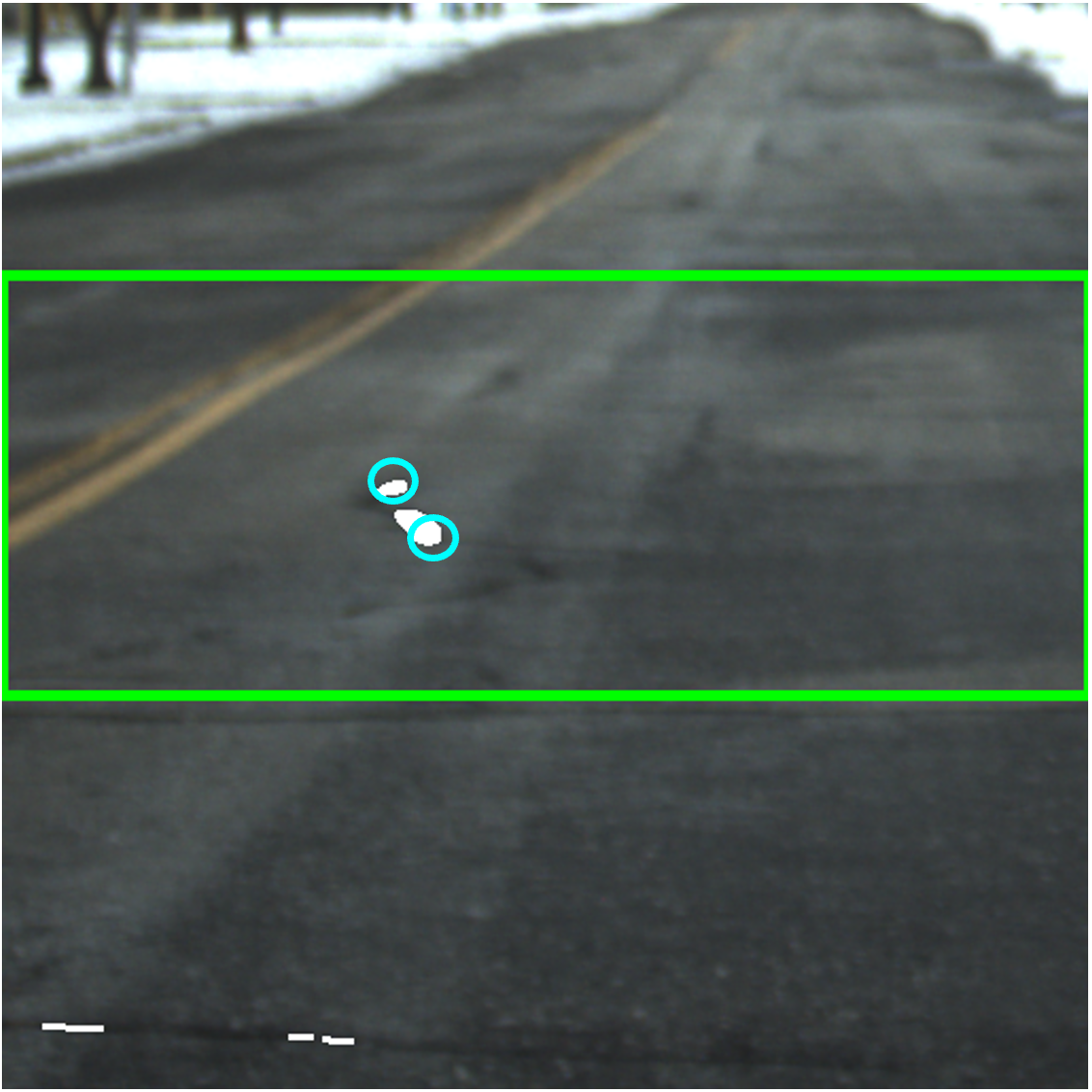}
& \includegraphics[width=2.1cm]{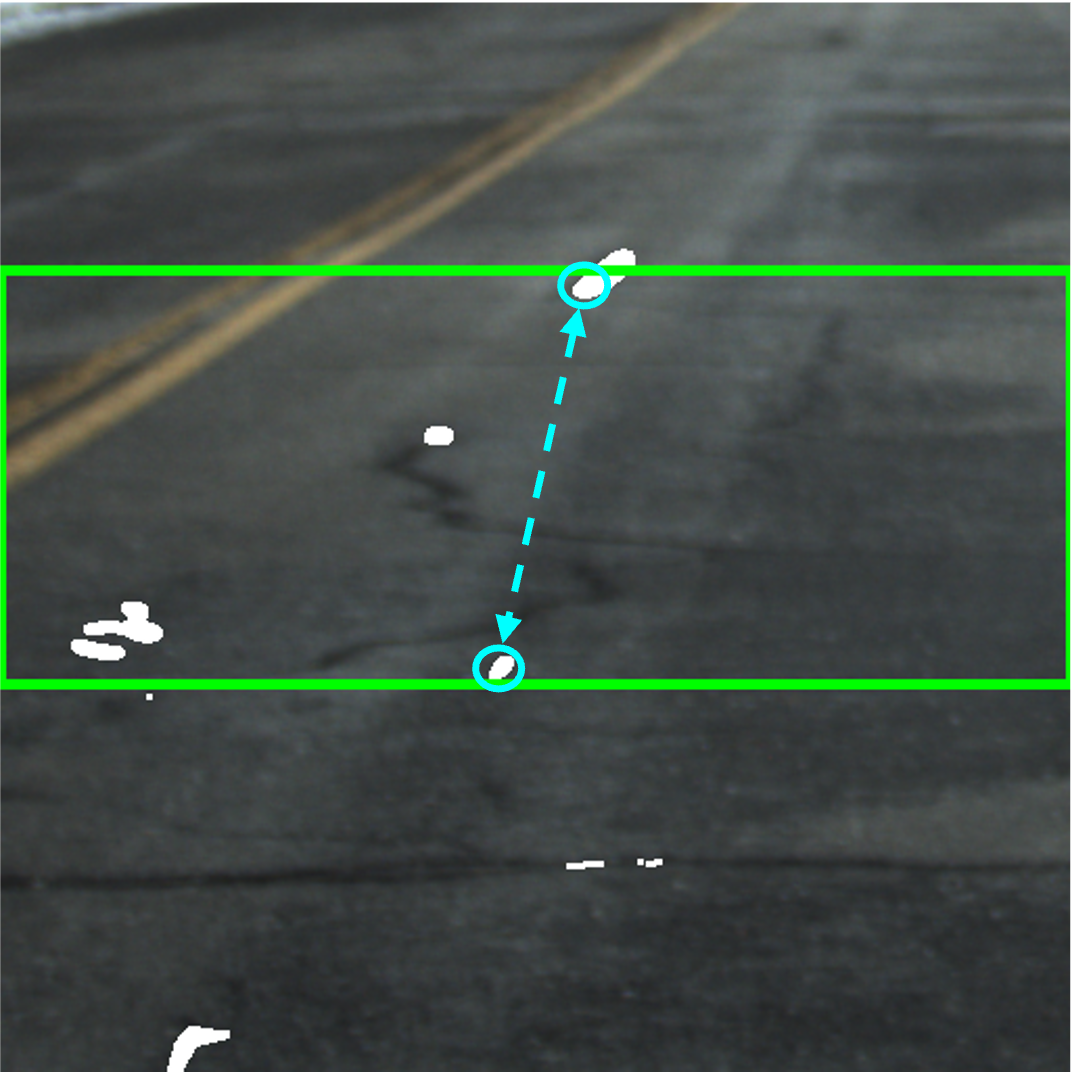}
& \includegraphics[width=2.1cm]{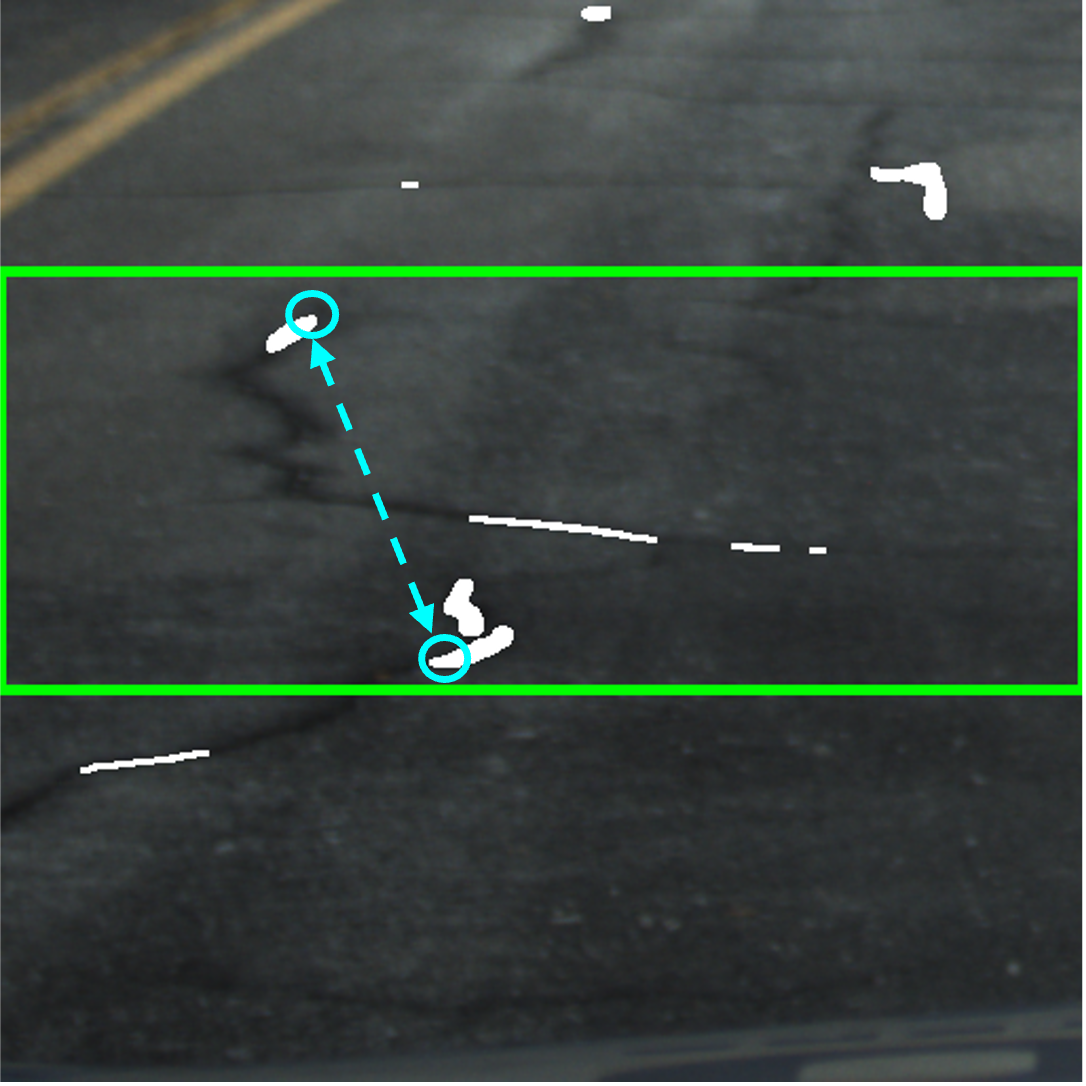}
& 1.4788 / 8.1916 / \textbf{3.3506} \\

Our Model w/ RT + Aug V3
& \includegraphics[width=2.4cm]{Figures/Target_crack.pdf}
& \includegraphics[width=2.1cm]{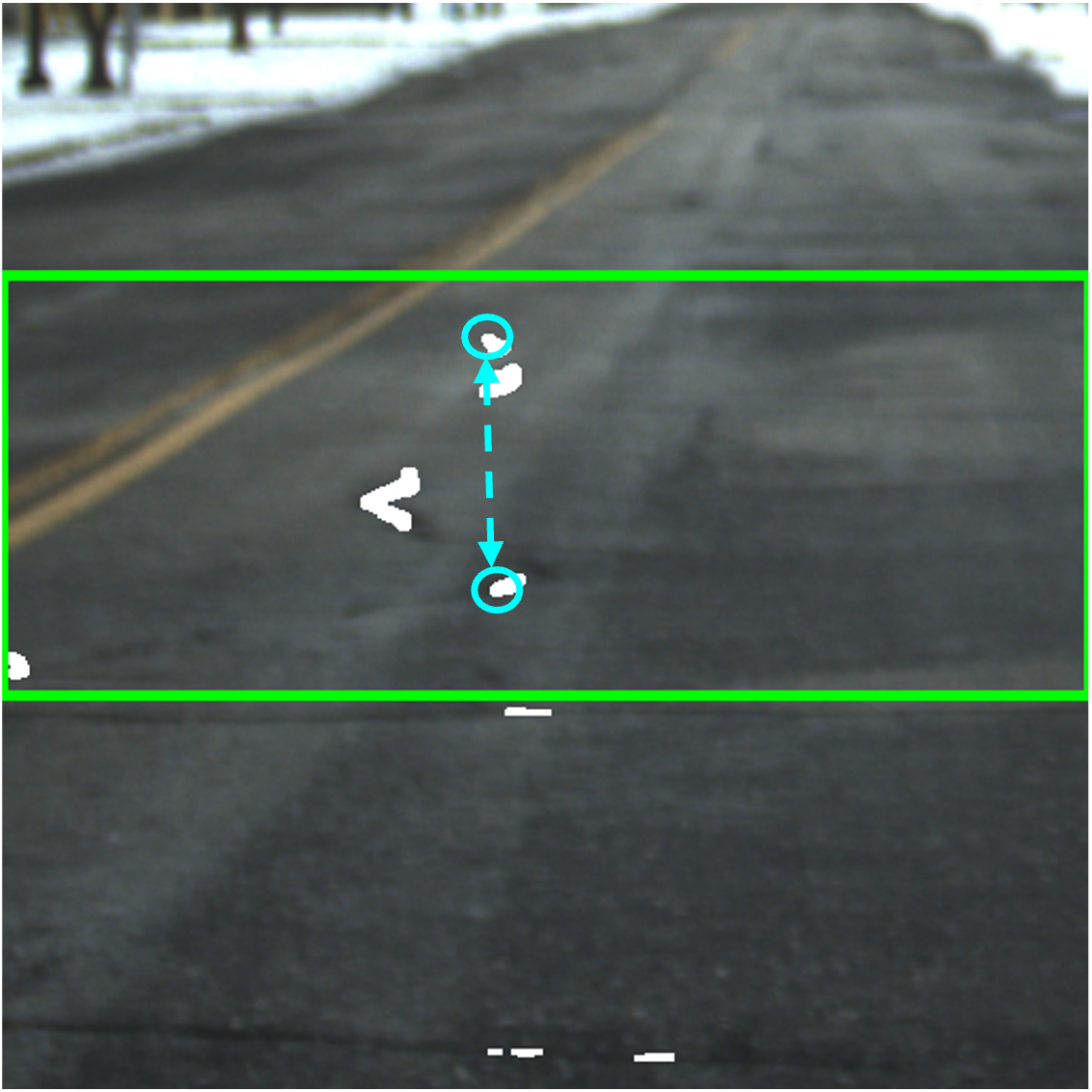}
& \includegraphics[width=2.1cm]{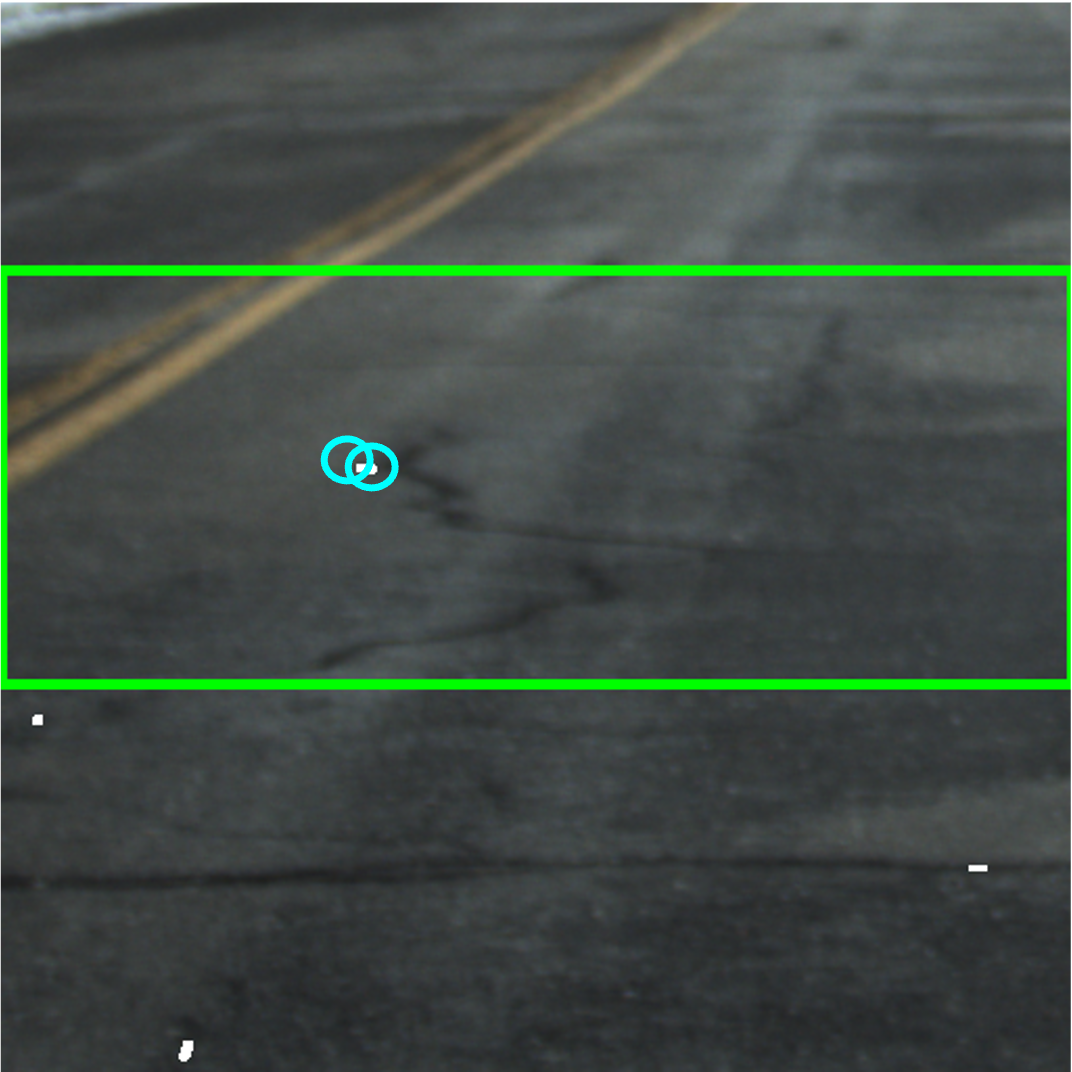}
& \includegraphics[width=2.1cm]{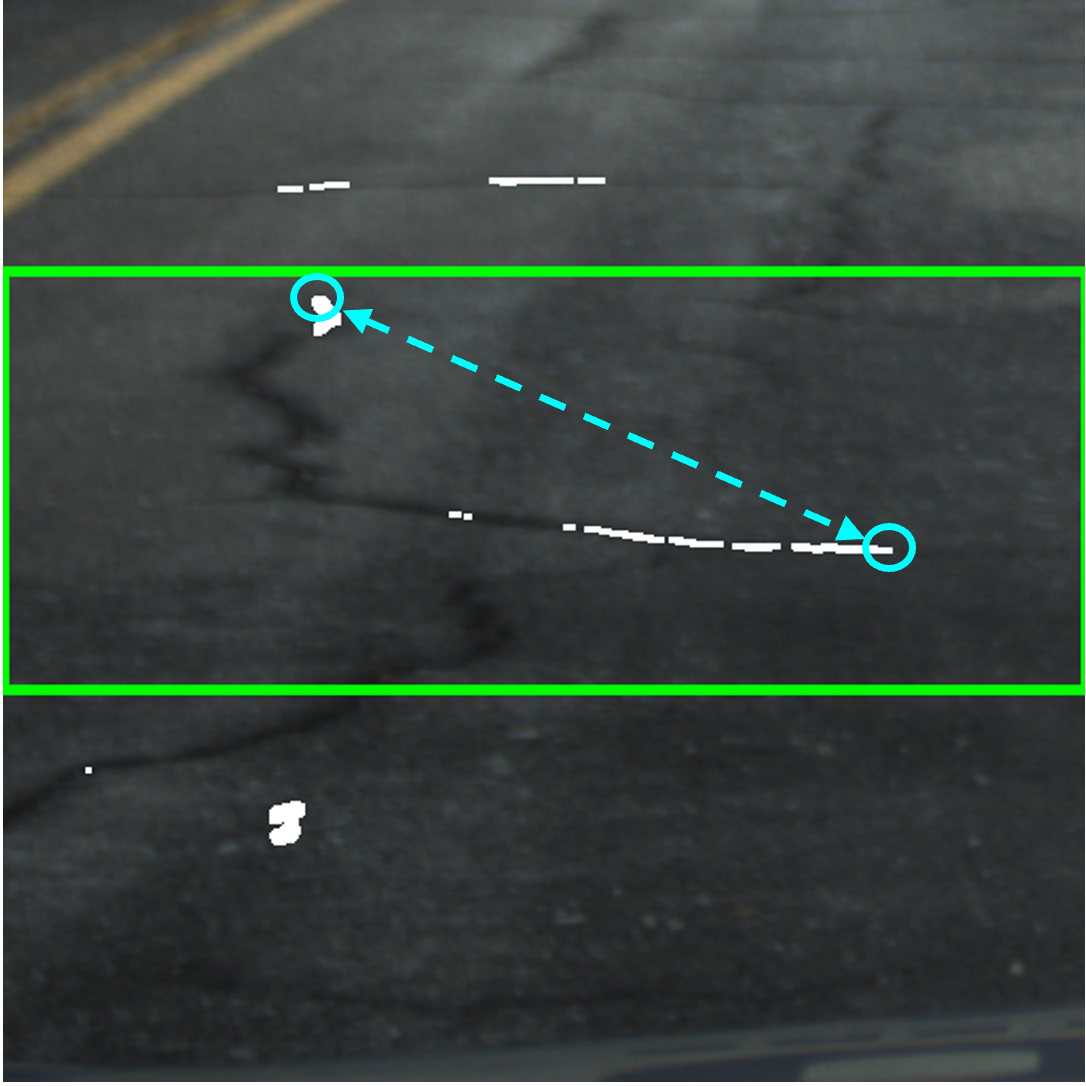}
& 7.7514 / 0.0865 / \textbf{2.9499} \\

\hline
\end{tabular}
\end{table*}
Once the range satisfies the resolution requirement for the ROI, the image subscription starts. 
To allow proper synchronization, raw images are logged with timestamp for downstream synchronization.
After passing the crack, the images are synced with other information (e.g., GPS) and the cropping windows for images are determined and saved along with each images.

\subsubsection{Stage $\mathbf{III}$: Post-detection}
\label{sssec: postd}
In this final stage, the detection node accesses the logged data directory and and extracts a 512$\times$512 window from each raw images. 
These cropped images will be passed on to the detection model trained in Section \ref{sec:crack_detection_model} to generate the detected masks. 
For each mask, the detection node identifies the edge coordinates using the method introduced in Section \ref{sec:length_estimation} and computes the Euclidean distance between edge points in the second and fourth quadrants as the estimated crack length.
The computed length results and the mask images are packaged and transmitted back to the RSU following the protocol in Fig.~\ref{fig:protocol_design}. 

\subsection{Ground truth validation and performance evaluation}\label{subsec:pipeline_performance}
To validate our pipeline's estimated crack length, we use a ZED 2i stereo camera with its software development kit \cite{stereolabs_zed2i}, which provides direct metric depth through stereo triangulation, to compute the crack length result using the same pinhole camera model described in \eqref{eqn:pinhole}. 
Results in Table.~\ref{tab:resolution_comparison} compares the crack length estimates at different resolutions of the center cropping using four different models. 
The ground truth corresponds the the red dashed line marked with ``Target crack" in the ``Target Crop" column. 
In the Mask columns, the edge corners determined using Algorithm \ref{algo:Edge_Selection} are marked as cyan dots and the corresponding cyan dashed line segments corresponds to the crack length estimation. 
Crops obtained when the vehicle is far from the crack result in lower effective resolution, as the crack occupies fewer pixels in the image, whereas crops obtained when the vehicle is closer to the crack provide higher effective resolution, with more pixels representing the crack.
The results show that models operating at high resolutions consistently outperform those at low resolutions, further emphasizing the importance of the motivation introduced in Section ~\ref{sec: motivation}. 
In addition, our trained model produces length estimates that are closer to those computed using the ZED 2i stereo depth, particularly as the crack-to-camera distance decreases. 
Moreover, the Baseline 2 model outperformed Baseline 1 in the area of the thin designated crack at high resolution, whereas both our models demonstrate superior detection of the crack's thin and low-contrast sections compared with both baselines. 

\section{Conclusion}\label{sec:conclusion}
In this paper, we designed an infrastructure-guided communication-enhanced road crack detection pipeline. 
Leveraging C-V2X communication between the infrastructure and the vehicle, the proposed framework dynamically localizes the region of interest within the vehicle's field of view, streamlines subsequent model inference, and estimates length. 
We investigated the performance of the state-of-the-art detection models on our carefully prepared real-time testing set (RT100). 
Our results demonstrate a 89\% improvement in the model performance when testing on front-view camera images. In addition, the estimated crack lengths, based on calibrated camera extrinsic parameters, achieve an accuracy of approximately 90\% compared to ground truth measurements.  

For future work, we will focus on improving the robustness of the proposed platform, particularly by evaluating its performance across diverse terrains and in scenarios involving multiple cracks within a single region of interest (ROI). 
Furthermore, we aim to develop an adaptive auto-focus mechanism that better localizes the region of interest to enhance dynamic cropping and further improve detection performance, integrating vehicle motion control \cite{xiao2025safe}.

\section*{Acknowledgment}
The authors acknowledge the financial support provided for this study by the Transportation Infrastructure Precast Innovation Center (TRANS-IPIC) through the University Transportation Center program of the US Department of Transportation, Office of the Assistant Secretary for Research and Technology (OST-R) under Grant No. 69A3552348333.

\bibliographystyle{unsrt}
\bibliography{bib}

\end{document}